\documentclass{article}

\usepackage{arxiv}

\usepackage[utf8]{inputenc} 
\usepackage[T1]{fontenc}    
\usepackage{hyperref}       
\usepackage{url}            
\usepackage{booktabs}       
\usepackage{amsfonts}       
\usepackage{nicefrac}       
\usepackage{microtype}      
\usepackage{lipsum}
\usepackage{graphicx}
\graphicspath{ {./images/} }
\usepackage{algorithm}
\usepackage{algpseudocode}
\usepackage{tabularx}
\usepackage{microtype}
\usepackage{array}
\usepackage{multirow}
\usepackage{booktabs}
\usepackage{graphicx}
\usepackage[table]{xcolor}
\usepackage[utf8]{inputenc}
\usepackage{xr-hyper}
\usepackage{xcolor}
\hypersetup{filecolor=red}
\usepackage{amsmath}
\usepackage{xspace}
\newcommand{\etal}{\textit{et al}.\xspace}

\title{Real-time Unsupervised Object Discovery from Asynchronous Event Streams}

\author{
 Pratham G. Shenwai \\
  School of Engineering and Technology\\
  University of New South Wales\\
  Canberra, Australia\\
  \texttt{p.shenwai@unsw.edu.au} \\
   \And
 Hemant Kumar Singh \\
  School of Engineering and Technology\\
  University of New South Wales\\
  Canberra, Australia\\
  \texttt{hemant.singh@unsw.edu.au} \\
  \And
  Sridhar Ravi \\
  School of Engineering and Technology\\
  University of New South Wales\\
  Canberra, Australia\\
  \texttt{sridhar.r@unsw.edu.au} \\
}

\begin{document}
\maketitle
\begin{abstract}
Event cameras capture pixel-level intensity changes with microsecond resolution to produce highly sparse asynchronous data streams. For visual perception in latency-critical environments, we propose a lightweight, training-free framework for discovery of moving objects based on spatio-temporal clustering. This framework is driven by two core contributions. First, a linear-time Spatio-temporal Probabilistic Event Filter (SPEF) that introduces an adaptive event acceptance threshold to distinguish salient motion structures from background noise. Second, an Event Morton Code Clustering (EMCC) module that bypasses expensive distance matrix computation to efficiently group events for unsupervised discovery of moving objects. On the E-MLB dataset benchmark, SPEF achieves the best denoising performance among classical filtering methods and remains competitive with learning-based approaches without requiring any offline training. On object discovery, EMCC achieves the highest overall accuracy and lowest execution time across the FRED and eTraM datasets, outperforming established density-based clustering baselines by a substantial margin. Overall, this work establishes a new performance benchmark for classical object discovery in event data, providing a highly scalable, training-free solution for resource-constrained visual perception. The code is available at \url{https://github.com/PrathamShenwai/SPEF_EMCC}
\end{abstract}

\section{Introduction}
\label{sec:intro}

Unsupervised object discovery from event camera streams is the task of identifying and localising coherent spatio-temporal event patterns, or blobs, corresponding to independently moving physical entities, without reliance on semantic labels or geometric object templates \cite{discovery1,discovery2,Discovery3,Discovery4, blob}. Closely related work frames this as moving object detection \cite{moving_obj,moving2,moving3,moving4,moving5}, and the two settings coincide wherever motion is the only available cue. Our setting differs in assuming neither predefined object categories nor supervised training. Discovery determines \emph{whether} a distinct moving entity exists and \emph{where} it manifests in the raw event stream, which makes it a prerequisite for downstream recognition or tracking in open-world settings \cite{prereq1,prereq2,prereq3}. Despite the event camera's favourable properties of microsecond temporal resolution and high dynamic range \cite{ASCNN,DAVIS,Evcam,Survey}, rapid and reliable unsupervised discovery from such streams remains an open problem.

Strong discovery requires solving two tightly coupled sub-problems: denoising and spatio-temporal clustering. Event streams are corrupted by structured noise arising from illumination flicker, background motion, and sensor hot pixels, all of which distort the neighborhood relationships that clustering depends on \cite{dwf,knoise,GMCM}. Unlike isolated outliers, this noise is spatio-temporally distributed and overlaps with true object motion, so filtering applied independently of motion structure is insufficient \cite{RobMohoney,E-MLB,noisy}. Standard denoisers optimise suppression accuracy in isolation, without regard for the event distributions that downstream clustering assumes.

\begin{figure}[!t]
  \centering
  \includegraphics[width=0.8\columnwidth]{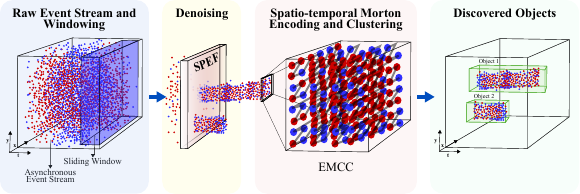}
  \caption{Overview of the proposed unsupervised event‑based object discovery framework. Events are processed in sliding temporal windows, denoised using the proposed \textbf{Spatio‑temporal Probabilistic Event Filter (SPEF)}, encoded with Morton codes to preserve spatio‑temporal locality, and clustered via \textbf{Event Morton Code Clustering (EMCC)} to produce 2D bounding box proposals.}
  \label{fig:top}
\end{figure}

The clustering stage poses a complementary bottleneck. Classical density-based methods such as DBSCAN \cite{DBSCAN1} and HDBSCAN \cite{HDBSCAN} operate without category priors and accommodate objects of varying density, but their repeated neighborhood queries in three-dimensional space-time incur costs that grow unfavorably with event count and neighborhood size \cite{prereq2,dwf,moving2}. This accuracy--efficiency trade-off is the central open challenge for real-time event-based object discovery.

We resolve this trade-off by aligning the denoising and clustering stages with the intrinsic geometry of the problem. Morton (Z-order) codes \cite{morton1966computer} map three-dimensional spatio-temporal coordinates to a one-dimensional scalar that preserves locality by construction. When the denoiser explicitly retains the neighborhood coherence that Morton ordering leverages, clustering reduces to \emph{gap detection in a sorted one-dimensional sequence}, eliminating the dominant computational bottleneck of classical clustering methods. Overview of our method is illustrated in Figure \ref{fig:top}.

The main contributions of this work are:
\begin{itemize}
    \item We introduce SPEF, a spatio-temporal probabilistic event filter that accepts each event with probability directly proportional to the product of a spatial activity score and a temporal coherence score. Unlike existing denoisers that impose a fixed  signal-noise decision threshold, SPEF derives its acceptance boundary directly from the live event distribution, requiring no manual calibration or learned model.
    \item We introduce EMCC, which reformulates event-stream clustering as gap detection on a Morton-ordered sequence, reducing clustering to an $O(N \log N)$ sort and $O(N)$ scan, eliminating neighborhood search entirely as a computational primitive for event-stream perception.
    \item The combined SPEF--EMCC framework is a fully unsupervised, training-free pipeline for event-based object discovery that operates in real time on resource-constrained hardware, requiring no optical flow estimation, iterative optimization, or category supervision.
\end{itemize}
\section{Related Work}

Supervised methods define the accuracy ceiling when category labels and large annotated datasets are available. They encode events into structured representations, including voxel grids \cite{voxel1,voxel2} and time surfaces \cite{timesurfaces1,timesurfaces2,timesurfaces3}, and apply
convolutional \cite{CNN1,ASCNN,CNN3}, graph-based \cite{AEGNN,EVGNN,GNNAutomotive}, and
recurrent \cite{RNN1,RNN2} architectures originally developed for frame-based vision.
Spiking neural networks \cite{SNN1,SNN2,SNN3} exploit native event sparsity through spike-driven computation, though accuracy on large benchmarks trails dense recurrent approaches. Among the latter, RED introduces ConvLSTM with multi-scale detection heads \cite{RED}, RVT combines local and dilated self-attention with stateful recurrence to achieve top results on Gen1 and 1~Mpx \cite{RVT,Gen1}, and state-space models replace recurrent layers for improved robustness across inference frequencies \cite{SSM}. All share two structural assumptions incompatible with open-world deployment: events
must be buffered into fixed temporal windows, and detectors recognise only
categories seen during training. Open-vocabulary detectors \cite{Owl-vit,ground_dino} address the second limitation in the frame-based domain. DEOE \cite{DEOE} is the closest event-based analogue, adding a disentangled objectness head to RVT to separate foreground classification from novel-object discovery, though it still requires offline training on domain-specific data. Our work makes neither assumption, producing bounding-box hypotheses without learned representations or category supervision.

When supervision is absent, the dominant alternative is scene decomposition via contrast maximisation \cite{Cmax1,Cmax2}, which aligns static background events under parametric flow models while leaving independently moving objects as structured residuals. This idea underpins pipelines ranging from per-event motion compensation \cite{stoffregen} to energy minimisation with graph cuts \cite{graph-cuts} and cascaded multi-model fitting \cite{cascaded}. Our work occupies a distinct point in this design space: rather than decomposing the scene through flow estimation, we directly cluster the denoised event stream into bounding-box hypotheses, a formulation that avoids iterative optimisation and is better suited to resource-constrained real-time platforms. Reliable clustering in turn depends critically on the quality of the preceding denoising stage.

Liu~\etal \cite{STCF} established that noise events lack spatio-temporal correlation with their neighbours and can be rejected by checking for a recent neighbour within a small spatial window. Subsequent work refines this for hardware via $O(N)$-space designs \cite{knoise} and improves high-noise performance through dual-window FIFO filtering \cite{dwf,Guo}. Learning-based methods such as EDnCNN \cite{EDnCNN} and EventZoom \cite{eventzoom} achieve higher suppression accuracy at substantially greater latency and resource cost. Critically, Shiba~\etal \cite{shiba} demonstrate that noise and motion are tightly coupled in raw event data, meaning that filtering decisions inevitably alter the spatio-temporal structure that downstream clustering depends on. Yet all existing denoisers are evaluated in isolation, leaving unaddressed how filter output should be structured for efficient spatial indexing downstream. The proposed SPEF is designed to close this gap by preserving the spatio-temporal neighborhood coherence that Morton-ordered clustering requires, rather than maximising standalone suppression accuracy.

The clustering stage has most commonly been addressed with density-based methods like DBSCAN~\cite{ST-DBSCAN} and HDBSCAN~\cite{mcinnes2017hdbscan} since they require no prior on cluster count and naturally accommodate objects of varying density. Both have seen wide adoption in cluster tracking and robot navigation \cite{DBSCAN2,DBSCAN3,DBSCAN4}, and neuromorphic variants have been mapped onto spiking hardware for ultra-low-power inference \cite{DBSCAN5}. On conventional embedded processors, however, the dominant cost is repeated neighborhood search.  Query complexity scales unfavourably with both event count and neighborhood size, and this bottleneck becomes acute in dense event regimes. 
Morton's Z-order curve \cite{morton1966computer,morton2}, employed in this work, offers a route out of this bottleneck. It maps multi-dimensional coordinates to a one-dimensional scalar that preserves spatial proximity, a property formalised for associative database search \cite{orenstein} and adopted broadly in spatial indexing and point-cloud processing \cite{pointcloud1,pointcloud2,pointcloud3}. In these applications, Morton ordering improves cache behaviour of existing index structures,  but neighborhood search itself remains. EMCC instead sorts denoised events by Morton code, which preserves spatio-temporal proximity by construction and reduces clustering to gap detection in the sorted sequence. To our knowledge, this is the first use of gap detection in Morton-ordered event sequences as a clustering mechanism for event-stream perception.
\section{Method}
\begin{figure*}[!t]
  \centering
  \includegraphics[width=\textwidth]{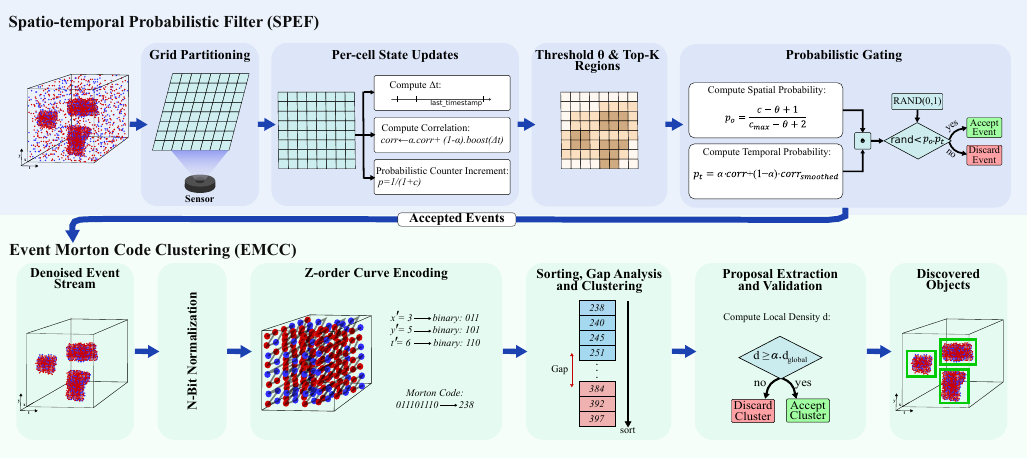}
  \caption{Overview of our object discovery framework using asynchronous event streams.
The framework begins with the Spatio-temporal Probabilistic Event Filter (SPEF), which maps events to grid cells and updates correlation scores using inter-event timing and probabilistic counters. Active regions are isolated via a top-k\% threshold, and events are stochastically filtered to maximize signal-to-noise while preserving structural integrity. The accepted events then flow into the Event Morton Code Clustering (EMCC) module. Here, they are normalized and assigned n-bit Morton (Z-order) codes to strictly preserve spatiotemporal locality. Sorting these codes allows for highly efficient clustering via gap detection in Morton space. Finally, candidate clusters are validated using geometric and statistical constraints to yield the final bounding box predictions.} 
  \label{fig:methodology}
\end{figure*}

Let $\mathcal{E} = \{e_k\}_{k=1}^{N}$, $e_k \doteq (x_k, y_k, t_k)$, denote the stream of events produced by an event camera of resolution $W \times H$ over a temporal window $[t_{\min}, t_{\max}]$. Unlike point clouds or image feature sets, $\mathcal{E}$ is asynchronous, sparse, and contaminated by noise events that are statistically indistinguishable from signal at the level of any individual $e_k$, so that structure is a collective property recoverable only through spatio-temporal grouping. This makes denoising and clustering mutually dependent: a filter that discards events without regard for local density structure corrupts the neighborhood relationships on which grouping relies, while a clusterer operating on unfiltered input conflates noise bursts with object boundaries. SPEF (Sec.~\ref{SPEF_Method}) breaks this dependence by accepting events in proportion to local spatio-temporal coherence, concentrating the filtered stream into the compact, object-level density regions that EMCC (Sec.~\ref{EMCC_Method}) encodes as Morton codes and partitions by gap detection on the resulting sorted sequence. The end-to-end pipeline is illustrated in Figure~\ref{fig:methodology}.

\subsection{Spatio-temporal Probabilistic Event Filter (SPEF)}
\label{SPEF_Method}

Event denoising methods typically impose fixed activity thresholds to separate signal from noise \cite{fixed_thresh1, fixed_thresh2}. A fixed threshold treats signal membership as a binary decision, which does not account for the continuous variation in spatio-temporal evidence, namely local activity and temporal coherence, across the sensor plane. Signal events and noise events differ not in any single measurable quantity but in the joint behaviour of spatial activity and temporal coherence \cite{STCF, knoise, EDnCNN}. Consider a region activated by a slow-moving object: the gradual change in brightness means the contrast threshold is exceeded infrequently, producing a modest event rate that may fall below a fixed activity threshold. Yet the events that are produced arrive with regular inter-event intervals, reflecting the steady motion of the object. A noise burst presents the opposite case: indoor lighting driven from AC mains produces intensity oscillations at twice the mains frequency, causing event cameras to generate periodic spurious events across the sensor plane \cite{RobMohoney} whose transient rate can exceed the same threshold, yet with no relationship to any moving object in the scene. A threshold on activity alone accepts the latter and rejects the former. No fixed operating point resolves both simultaneously. The problem is therefore not one of finding a better threshold, but of replacing the binary decision with one that is proportional to the evidence that each event belongs to the signal distribution.

 Each incoming event is a \textit{proposal} that may arise from a moving edge (signal) or sensor noise (background) \cite{Survey}, and no deterministic rule cleanly separates the two at the level of any individual event \cite{EDnCNN}. The question is therefore not \textit{``is this event signal?''} but \textit{``with what probability does this event belong to the 
signal distribution?''}

We implement this idea as a probabilistic event filter, with rejection sampling as the motivating intuition \cite{rejectionsampling}. No target distribution is specified in advance. The acceptance boundary is instead derived from the live stream through two evidence terms, a spatial term $p_o$ and a temporal term $p_t$, whose product governs acceptance. The product is jointly high only when a region exhibits both sustained activity and temporal coherence, the signature of a true signal event, producing a dynamic acceptance landscape that no fixed threshold can replicate.

\noindent\textbf{Grid partition and per-region state.}
To evaluate $p_o$ and $p_t$ efficiently, the sensor plane is partitioned into $R \times R$-pixel regions. Each event $e_k = (x_k, y_k, t_k)$ is mapped in $O(1)$ to its containing region:
\begin{equation}
    i_x = \lfloor x_k / R \rfloor, \quad i_y = \lfloor y_k / R \rfloor.
    \label{eq:ix-iy-floor}
\end{equation}
Per region, an event counter $c$ accumulates activity and serves as the basis for $p_o$.

\noindent\textbf{Spatial activity score $p_o$ and probabilistic counting.}
To prevent persistently active regions from saturating their counters, $c$ is incremented stochastically with probability
\begin{equation}
    p = \frac{1}{1+c},
    \label{eq:prob-increment}
\end{equation}
capped at $c_{\max} = 2^b - 1$ for a $b$-bit counter, preserving dynamic range across the full sensor plane. Sec.~\ref{Ablation} compares probabilistic and linear incrementing at two bit depths. An adaptive threshold $\theta$ is computed as the top-$k\%$ partial selection over all 
region counters (sensitivity to $k$ is analysed in Sec.~\ref{Ablation}). A region whose counter barely exceeds $\theta$ offers weak evidence of sustained activity. A region approaching $c_{\max}$ offers strong evidence. The spatial acceptance probability 
encodes this directly:
\begin{equation}
    p_o = \frac{c - \theta + 1}{c_{\max} - \theta + 2},
    \label{eq:po}
\end{equation}
satisfies $p_o \rightarrow 0$ as $c \rightarrow \theta$ and 
$p_o \rightarrow 1$ as $c \rightarrow c_{\max}$.

\noindent\textbf{Temporal coherence score $p_t$.}
To capture how regularly a region is being activated, each region additionally maintains a correlation score $corr[i_y,i_x]$, which accumulates evidence of temporal regularity by tracking the recency and consistency of event arrivals within that region. This score is updated on every incoming event via exponential smoothing driven by the elapsed time since the previous event in the same region:
\begin{equation}
    \text{corr}[i_y,i_x] \leftarrow
        \alpha \cdot \text{corr}[i_y,i_x] + (1-\alpha) \cdot \kappa\exp(-\Delta t / \tau),
    \label{eq:corr_update}
\end{equation}
where $\Delta t = t_k - \text{last\_timestamp}[i_y, i_x]$, $\alpha \in [0,1]$ controls 
temporal memory, and $\kappa$, $\tau$ set the amplitude and timescale. Rapidly recurring 
events, characteristic of a moving edge, receive high weight while sporadic noise 
receives near-zero weight. A $3 \times 3$ uniform box filter smooths the correlation 
matrix to broaden the coherence response around object regions, giving
\begin{equation}
    p_t = \alpha \cdot \text{corr}[i] + (1-\alpha) \cdot 
    \operatorname{uniform\_filter}(\text{corr},\,3)[i].
    \label{eq:pt}
\end{equation}

\noindent\textbf{Acceptance criterion.}
An event is accepted if and only if
\begin{equation}
    p_{\text{keep}} = p_o \cdot p_t,\quad r < p_{\text{keep}}, \quad r \sim \mathcal{U}(0,1).
    \label{eq:acceptance}
\end{equation}
A noise region may be transiently active ($p_o$ elevated, $p_t \approx 0$) or locally coherent but globally inactive ($p_t$ elevated, $p_o \approx 0$). The product suppresses both failure modes. Events at object peripheries survive with probability proportional to their temporal coherence rather than being discarded by a threshold they narrowly miss, preserving the boundary structure that downstream processing for object discovery requires. 

Since $r \sim \mathcal{U}(0,1)$, the fundamental property of the uniform distribution gives $\Pr(r < p_{\text{keep}}) = p_{\text{keep}}$, so each event is accepted with probability exactly equal to the strength of its spatio-temporal evidence. Because $\theta$ is derived from the live counter distribution rather than set globally, its numerical value shifts naturally with scene density: rising when the stream is dense and falling when it is sparse, without any manual recalibration. To our knowledge, SPEF is the first event denoising filter to accept each event with probability equal to its spatio-temporal evidence rather than at a fixed operating point. Section~S5 of the supplementary material replaces this criterion with fixed thresholds on $p_o \cdot p_t$ between 0.4 and 0.8, across three FRED scenarios covering multiple objects, a sparse distant target, and dynamic approach and recession. Probabilistic acceptance holds the highest F1 in all three, with the largest margin on the sparse target, while every fixed threshold degrades on the conditions it was not tuned for.

\subsection{Event Morton Code Clustering (EMCC)}
\label{EMCC_Method}
Clustering asynchronous event streams requires grouping events by spatio-temporal proximity. Existing methods address this through neighborhood graphs \cite{graph-cuts, GSCEventMOD}, kernel density estimates \cite{barranco}, or iterative centroid updates \cite{MondalKmeans}, all of which scale with the number of pairwise event relationships. EMCC takes a different approach: rather than computing proximity, it encodes proximity into a scalar ordering, reducing clustering to gap detection on a sorted sequence.

Morton codes \cite{morton1966computer} provide this encoding. By interleaving the binary representations of each coordinate dimension, a Morton code maps a multi-dimensional point to a scalar such that points close in the original space map to nearby scalars \cite{z-order1,z-order2,z-order3}. Sorting events by their Morton code linearizes the spatio-temporal cloud while preserving its local structure. Object boundaries appear as large gaps in the sorted sequence, while events from the same object produce small gaps. The result is a clustering procedure that avoids pairwise proximity computation entirely. While hardware-accelerated ray tracing \cite{karras2012maximizing} demonstrates the efficiency of sorted-list operations over locality-preserving codes, it operates on static geometry with fixed coordinates. Applying this principle to event streams is non-trivial: the spatio-temporal distribution of events varies continuously with scene dynamics, coordinates are not fixed at encoding time, and the gap structure in the sorted sequence must reflect live activity rather than a pre-built scene. EMCC addresses these challenges by deriving the partition threshold $\tau = \text{percentile}(\Delta, p_{\text{gap}})$ directly from the observed gap distribution, allowing the clustering to adapt to the density and structure of the incoming stream without requiring an explicit motion model or scene representation. The sensitivity to $p_{\text{gap}}$ is characterized in Sec.~\ref{Ablation}. To our knowledge, EMCC is the first method to apply Morton encoding to event-based object discovery.

EMCC processes events through four stages, each motivated by the asynchronous structure of event data. Since spatial coordinates (pixels) and temporal coordinates (microseconds) occupy incommensurable numerical ranges, both are normalised to a common $n$-bit integer space before Morton encoding, ensuring no single dimension dominates the interleaved code. The normalised events are encoded as Morton codes and sorted to produce a linearised sequence, in which gaps between consecutive codes reflect spatio-temporal separation. These gaps partition the sequence into clusters, though gap magnitude alone does not distinguish coherent objects from low-density regions, so each candidate cluster $\mathcal{C}_k$ is validated against the criterion $|\mathcal{C}_k| / A \geq \alpha \cdot d_{\text{global}}$, where $\alpha$ controls the strictness of the density requirement relative to the global event rate $d_{\text{global}} = N / (W \times H)$. Finally, the hierarchical structure of the Z-order curve means objects spanning a major bit boundary appear as adjacent sub-clusters rather than a single cluster, a known property of space-filling 
curves \cite{z-order1,z-order2}, so proposals whose centres fall within proximity threshold $d_{\text{link}}$ are merged before Non-Maximum Suppression yields the final detection set $\mathcal{B}$. The complete procedure is summarised in Algorithm~\ref{alg:EMCC}, with full implementation details provided in Algorithm~SA1 of the supplementary material.

\noindent\textbf{Normalisation and encoding.} Each event $e_k = (x_k, y_k, t_k)$ is normalised 
as follows:
\begin{equation}
    x' = \left\lfloor \frac{x}{W}(2^n-1) \right\rfloor, \quad
    y' = \left\lfloor \frac{y}{H}(2^n-1) \right\rfloor, \quad
    t' = \left\lfloor \frac{t-t_{\min}}{t_{\max}-t_{\min}}(2^n-1) \right\rfloor.
    \label{eq:normalisation}
\end{equation}
Normalised coordinates are encoded as $m_k = \text{MortonEncode}(x', y', t')$ and sorted to produce the linearised sequence $\sigma = \text{argsort}(m)$.

\noindent\textbf{Gap-based clustering.} Consecutive gaps in the sorted sequence,
\begin{equation}
    \Delta_i = m_{\sigma(i+1)} - m_{\sigma(i)}, \quad i = 1, \ldots, N-1,
    \label{eq:gaps}
\end{equation}
are thresholded at $\tau = \text{percentile}(\Delta, p_{\text{gap}})$ to partition the sequence into clusters $\mathcal{C} = \{\mathcal{C}_1, \mathcal{C}_2, \ldots\}$ wherever $\Delta_i > \tau$.

\noindent\textbf{Density-based validation.} For each cluster $\mathcal{C}_k$, a bounding box is extracted using spatial percentiles $(p_{\text{low}}, p_{\text{high}})$, yielding area $A$. The cluster is retained only if
\begin{equation}
    \frac{|\mathcal{C}_k|}{A} \geq \alpha_d \cdot d_{\text{global}}.
    \label{eq:density}
\end{equation}
where $\alpha_d$ is the density multiplier.
\noindent\textbf{Spatial aggregation.} Box proposals whose centres fall within $d_{\text{link}}$ are merged, and Non-Maximum Suppression yields the final detection set $\mathcal{B}$. Sensitivity to $n$, $p_{\text{gap}}$, $\alpha_d$, $d_{\text{link}}$ is characterised in Sec.~\ref{Ablation}. 

\begin{algorithm}[t]
\caption{Event Morton Code Clustering (EMCC)}
\label{alg:EMCC}
\begin{algorithmic}[1]
\Require Events $\mathcal{E} = \{(x_i, y_i, t_i)\}_{i=1}^{N}$, resolution $W \times H$
\Ensure Bounding boxes $\mathcal{B}$
\State $d_{\text{global}} \gets N / (W \times H)$
\State Normalise each event to $n$-bit space per Eq.~\ref{eq:normalisation}
\State $m_k \gets \text{MortonEncode}(x'_k, y'_k, t'_k)$ for each event
\State $\sigma \gets \text{argsort}(m)$
\State Compute gaps $\Delta_i$ per Eq.~\ref{eq:gaps} and partition into 
       $\mathcal{C}$ where $\Delta_i > \tau$
\State For each $\mathcal{C}_k$, extract bounding box via 
       $(p_{\text{low}}, p_{\text{high}})$
\State Retain $\mathcal{C}_k$ satisfying Eq.~\ref{eq:density}
\State Merge proposals within $d_{\text{link}}$, apply NMS 
       $\rightarrow \mathcal{B}$
\end{algorithmic}
\end{algorithm}
\section{Experiments}
\label{Experiments}
\noindent\textbf{Datasets:} A typical event-based perception framework comprises ego-motion estimation, compensation for ego-motion, denoising, and clustering. Our proposed methods, SPEF and EMCC, target the final two stages and thus assume motion-compensated input for performance evaluations here. This assumption is supported on two grounds. First, both discovery benchmarks use statically mounted sensors, so Table~\ref{tab:clustering_comparison} already reports the zero-ego-motion regime representative of widespread applications such as surveillance and traffic monitoring. Second, every operating point in the pipeline is defined on the live input: the spatial gate $\theta$ tracks the current counter distribution, the gap threshold is a percentile of the observed gaps, and cluster retention scales with the live event rate $d_{\text{global}}$. Residual compensation error shifts the input statistics without destroying spatio-temporal structure, so degradation is expected to be gradual, manifesting as the merging of nearby objects rather than abrupt failure. The modular interface is compatible with common upstream compensators, including IMU warping, contrast maximisation, and learned compensation \cite{zhao, Survey}. As a first step, to isolate and comprehensively validate the denoising performance of SPEF, we test it on the E-MLB dataset \cite{emlb} across diverse day and night sequences, using the Mean Event Structural Ratio (MESR) metric to quantify performance. Subsequently, for the EMCC event clustering module, we evaluate performance on two distinct benchmarks: the Florence RGB-Event Drone Dataset (FRED) \cite{magrini2025fred} and the eTraM traffic dataset \cite{etram}. These two datasets were chosen because they are well established and contain samples with large variation in ambient conditions, event density and object salience. 

\noindent\textbf{Hardware Setup:} 
All experiments were conducted on a laptop equipped with a 13th Gen Intel Core i7-1360P CPU~(12 cores, 16 threads) and 32 GB RAM. This modest setup highlights the computational efficiency of the proposed method, which achieves real-time performance using only CPU resources. \\

\noindent\textbf{Baselines.}
EMCC is benchmarked against three unsupervised clustering methods: ST-DBSCAN \cite{ST-DBSCAN}, ST-MeanShift \cite{meanshift}, and ST-HDBSCAN \cite{mcinnes2017hdbscan}, where 'ST' stands for spatio-temporal versions of these algorithms. These represent the three 
foundational paradigms of density-based spatial grouping: fixed-radius neighborhood analysis, kernel density estimation, and hierarchical density decomposition. Learning-based and graph-based methods are excluded from this comparison as they operate under a fundamentally different problem formulation, requiring labelled training data and task-specific supervision, which places them outside the zero-shot scope of this evaluation. 
The comparison is therefore restricted to methods that, like EMCC, require no learned priors and operate directly on raw event streams. \\

\noindent\textbf{Evaluation protocol.}
Event clusters produced by all methods are converted to bounding box proposals using the same geometric bounding procedure, ensuring that differences in metric scores reflect clustering quality rather than proposal extraction. Precision, Recall, F1, and IoU are reported against ground truth annotations within a 33\,ms sliding window, matching the ground truth annotation interval of the evaluation datasets. Hyperparameters for all methods are tuned on validation sets drawn from the training splits of these datasets, sized at 25\% of the number of test sequences (12 sequences for FRED, 10 day/night sequences for eTraM). Algorithm-specific parameters and shared post-processing parameters are optimised jointly to prevent configuration bias toward any single method. Tuning ranges are detailed in the supplementary material.

\subsection{Comparison with Baselines}
\begin{figure}[!t]
  \centering
  \includegraphics[width=\columnwidth]{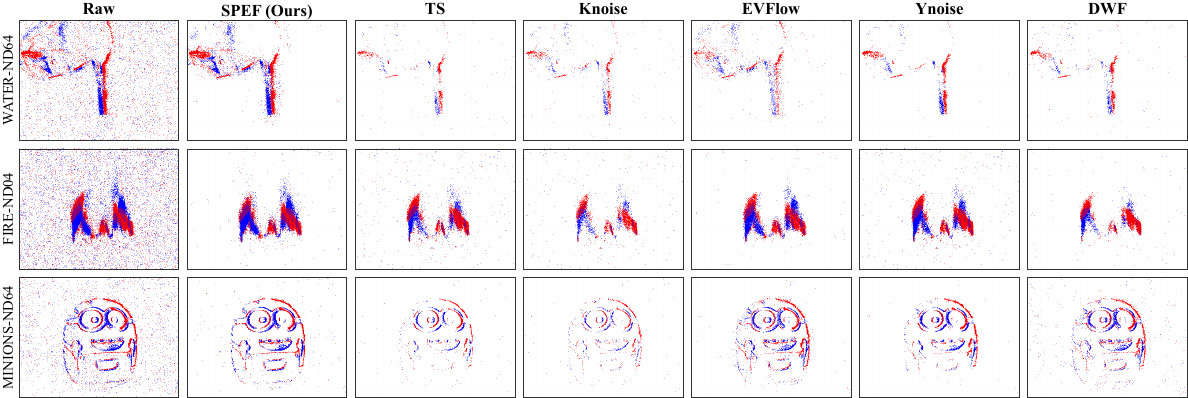}
  \caption{Comparison of denoising methods on three E-MLB \cite{emlb} sequences across ND04 and ND64 noise levels}
  \label{fig:filter_qualitative}
\end{figure}

For evaluation of the filtering stage, we benchmarked SPEF over the large-scale E-MLB dataset \cite{emlb}, which has been used by a number of previous denoising studies \cite{edformer,emlb,simultaneous,Ding}. For  quantitative evaluation of the filter performance, the Mean Event Structural Ratio (MESR$\uparrow$) is used. MESR quantitatively evaluates the preservation of the essential spatiotemporal contiguity while eliminating uncorrelated background noise. We adopt the denoising benchmarking framework from  Shiba et al. \cite{simultaneous} to evaluate filter performance. Specifically, we expand their published results by appending the SPEF metrics to their comparative table, classifying the filter as a model-based approach. Qualitative denoising outputs across representative day and night sequences are shown in Figure~\ref{fig:filter_qualitative}.

\begin{table}[t]
\centering
\caption{Comparison of event denoising methods using the MESR$\uparrow$ metric on E-MLB \cite{emlb} dataset. Best results are \textbf{bold}, and second best are \underline{underlined}.}
\label{tab:denoising}
\scriptsize 
\setlength{\tabcolsep}{1.2pt} 
\renewcommand{\arraystretch}{0.95} 
\begin{tabular}{clcccccccc} 
\toprule
& \multirow{2}{*}{Method} & \multicolumn{4}{c}{E-MLB (Day)} & \multicolumn{4}{c}{E-MLB (Night)} \\ 
\cmidrule(lr){3-6} \cmidrule(lr){7-10}
& & ND1 & ND4 & ND16 & ND64 & ND1 & ND4 & ND16 & ND64 \\ 
\midrule
\multirow{11}{*}{\rotatebox{90}{Model-based}} 
& Raw           & 0.821 & 0.824 & 0.815 & 0.786 & 0.890 & 0.824 & 0.786 & 0.768 \\
& BAF \cite{BAF}     & 0.861 & 0.869 & 0.876 & 0.890 & 0.946 & \underline{0.973} & \underline{0.992} & \underline{0.942} \\
& TS \cite{Lagorce}     & 0.877 & 0.887 & 0.870 & 0.837 & 1.033 & 0.944 & 0.886 & 0.797 \\
& KNoise\cite{knoise} & 0.846 & 0.837 & 0.830 & 0.807 & 0.954 & 0.956 & 0.871 & 0.817 \\
& EvFlow \cite{evgait} & 0.848 & 0.878 & 0.868 & 0.833 & 0.969 & \textbf{0.983} & 0.889 & 0.797 \\
& IETS \cite{Baldwin}    & 0.772 & 0.785 & 0.777 & 0.753 & 0.950 & 0.823 & 0.804 & 0.711 \\
& Ynoise \cite{ynoise} & 0.866 & 0.863 & 0.857 & 0.821 & 1.009 & 0.943 & 0.875 & 0.792 \\
& GEF \cite{GEF}    & \textbf{1.051} & 0.938 & 0.935 & 0.927 & \underline{1.027} & 0.955 & 0.946 & 0.935 \\
& DWF \cite{dwf}    & 0.878 & 0.876 & 0.866 & 0.865 & 0.923 & 0.962 & 0.988 & 0.932 \\
& ESMD \cite{simultaneous} & 0.938 & \textbf{0.958} & \underline{0.986} & \textbf{0.950} & \textbf{1.037} & 0.961 & 0.945 & 0.932 \\
& \textbf{SPEF (Ours)}  & \underline{0.974} & \underline{0.944} & \textbf{1.004} & \underline{0.938} & 1.023 & 0.964 & \textbf{1.005} & \textbf{1.051} \\
\midrule
\multirow{4}{*}{\rotatebox{90}{Learning}}
& EDnCNN \cite{EDnCNN}  & 0.887 & 0.908 & 0.903 & 0.912 & 1.001 & \textbf{1.024} & \textbf{1.079} & \underline{1.086} \\
& EventZoom \cite{eventzoom}& \textbf{0.996} & \textbf{0.988} & \textbf{0.996} & \textbf{0.970} & \textbf{1.055} & 1.007 & 1.010 & 0.988 \\
& MLPF \cite{dwf}   & 0.851 & 0.855 & 0.846 & 0.840 & 0.926 & 0.928 & 0.910 & 0.906 \\
& EDformer \cite{edformer}& \underline{0.952} & \underline{0.955} & \underline{0.956} & \underline{0.942} & \underline{1.048} & \underline{1.019} & \underline{1.076} & \textbf{1.099} \\
\bottomrule
\end{tabular}
\end{table}

As shown in Table \ref{tab:denoising}, SPEF demonstrates highly competitive performance among the model-based methods, particularly under severe noise conditions. Specifically, SPEF achieves the highest MESR scores at the ND16 level for daytime sequences and takes a clear lead at the highest noise thresholds (ND16 and ND64) in challenging nighttime environments. It also performs second-best in Daytime ND1, ND4 and ND64 test cases. Beyond outperforming other model-based filters, SPEF substantially narrows the performance gap with recent learning-based architectures. Despite using probabilistic thresholds, the method yields structure preservation metrics that rival, and occasionally surpass, those of state-of-the-art supervised networks. Importantly, SPEF eliminates uncorrelated background scatter while maintaining the structural integrity of moving targets. The resulting sparse event stream significantly reduces the downstream search space, providing a base for the spatio-temporal grouping. The performance demonstrated here establishes SPEF as a robust and practical solution for real-world event denoising across varied conditions.

\begin{figure*}[!t]
  \centering
  \includegraphics[width=\textwidth]{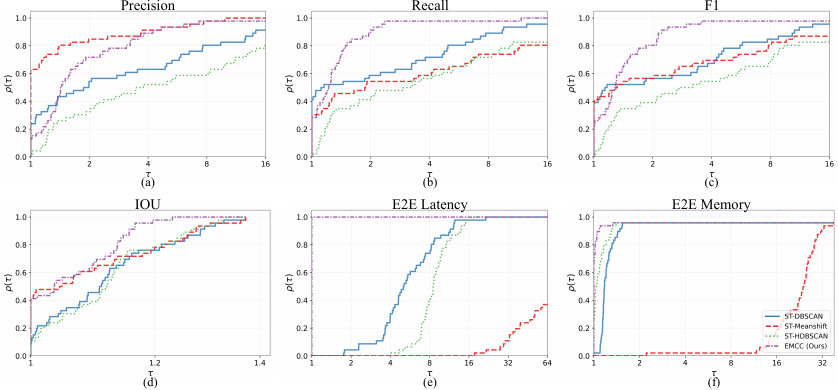}
  \caption{Performance profiles of four clustering algorithms evaluated on the FRED \cite{magrini2025fred} dataset. The curves plot the performance ratio threshold ($\tau$) against the fraction of sequences ($\rho(\tau)$) where an algorithm's metric is within a factor of $\tau$ of the top-performing method. Steeper curves that reach $\rho(\tau)=1.0$ at lower $\tau$ values indicate greater performance consistency across the dataset. As shown in the profiles, ST-MeanShift (red dashed) achieves the highest initial Precision at $\tau=1$. However, our proposed EMCC (purple dash-dotted) exhibits robust overall performance, reaching $\rho(\tau)=1.0$ faster than ST-DBSCAN, ST-MeanShift, and ST-HDBSCAN in Recall, F1, and IOU. Notably, EMCC achieves the best relative performance ($\rho(\tau)=1.0$) for E2E Latency across all sequences at $\tau=1$ and shows significantly higher consistency in E2E Memory compared to the baselines. Profiles shown: (a) Precision, (b) Recall, (c) F1, (d) IOU, (e) Mean E2E Latency, and (f) Mean E2E Memory.}
  \label{fig:wide_figure_4}
\end{figure*}

\begin{figure}[!bp]
  \centering
  \includegraphics[width=\columnwidth]{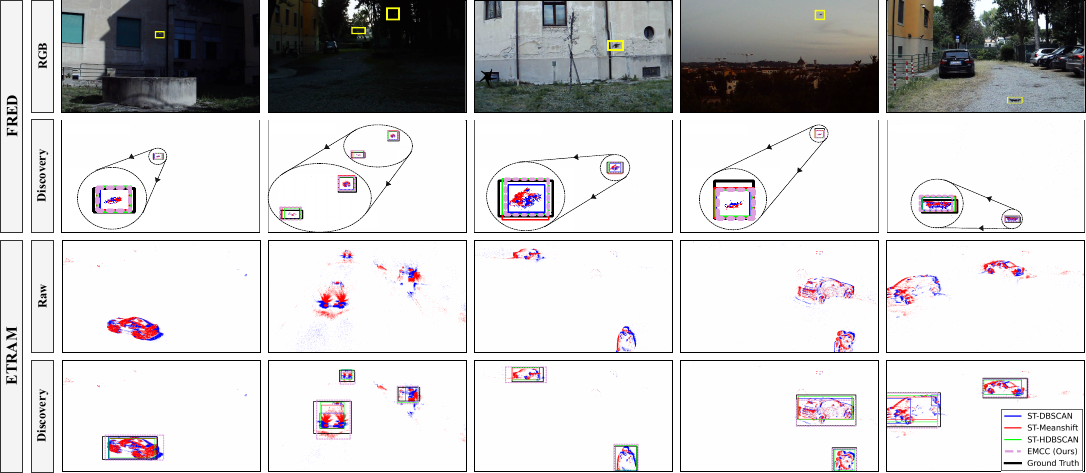}
  \caption{Qualitative comparison of object discovery on FRED \cite{magrini2025fred} and eTraM \cite{etram} datasets. For FRED, yellow boxes on RGB frames provide visual reference, with black-dashed insets showing zoomed-in views of discovered objects. Bounding box proposals are displayed for ST-DBSCAN (blue), ST-MeanShift (red), ST-HDBSCAN (green), EMCC (purple-dashed), and Ground Truth (black).}
  \label{fig:EMCC_qualitative}
\end{figure}

\begin{table}[!t]
\centering
\caption{Unsupervised Spatio-temporal Object Discovery Performance on FRED and eTraM. All methods process SPEF-filtered event streams. ST denotes Spatio-temporal and indicates our $x,y,t$ implementations of the referenced algorithms. Latency is reported in ms. Best results are \textbf{bold}, and second best are \underline{underlined}.}
\label{tab:clustering_comparison}
\scriptsize 
\setlength{\tabcolsep}{5pt} 
\renewcommand{\arraystretch}{0.95} 
\begin{tabular}{lccccc}
\toprule
\textbf{Method} & \textbf{P$\uparrow$} & \textbf{R$\uparrow$} & \textbf{F1$\uparrow$} & \textbf{IOU@50$\uparrow$} & \textbf{E2E Lat. (ms)$\downarrow$} \\
\midrule
\multicolumn{6}{c}{\textit{FRED \cite{magrini2025fred}}} \\
\midrule
ST-DBSCAN \cite{ST-DBSCAN}        & 0.348 & \underline{0.402} & \underline{0.358} & 0.594 & \underline{27.3} \\
ST-Mean Shift \cite{meanshift} & \textbf{0.511} & 0.260 & 0.301 & \underline{0.631} & 517.1 \\
ST-HDBSCAN \cite{mcinnes2017hdbscan}      & 0.239 & 0.281 & 0.248 & 0.602 & 70.4 \\
\textbf{EMCC (Ours)}           & \underline{0.416} & \textbf{0.517} & \textbf{0.441} & \textbf{0.647} & \textbf{6.3} \\
\midrule
\multicolumn{6}{c}{\textit{eTraM \cite{etram}}} \\
\midrule
ST-DBSCAN \cite{ST-DBSCAN}        & 0.319 & \textbf{0.262} & \textbf{0.273} & \underline{0.674} & \underline{36.7} \\
ST-Mean Shift \cite{meanshift} & \underline{0.375} & 0.151 & 0.204 & 0.634 & 490.5 \\
ST-HDBSCAN \cite{mcinnes2017hdbscan}      & 0.332 & 0.245 & \underline{0.266} & \textbf{0.675} & 150.1 \\
\textbf{EMCC (Ours)}           & \textbf{0.378} & \underline{0.246} & \textbf{0.273} & 0.656 & \textbf{15.4} \\
\bottomrule
\end{tabular}
\end{table}

Building on the denoised output of SPEF, we evaluate our clustering algorithm, EMCC, on the FRED \cite{magrini2025fred} and eTraM \cite{etram} datasets. To ensure identical input quality, every tested method processes the exact same SPEF-filtered event stream. We compare our approach against three continuous spatio-temporal baselines: ST-DBSCAN \cite{ST-DBSCAN}, ST-MeanShift \cite{meanshift}, and ST-HDBSCAN \cite{mcinnes2017hdbscan}. We normalize event coordinates $(x,y,t)$ into a continuous unit cube for these baselines. This step is mandatory for fairness as it prevents microsecond timestamps from skewing spatial coordinates during euclidean distance calculations. 

Like the baseline clustering approaches, EMCC processes continuous sliding windows. Table \ref{tab:clustering_comparison} summarizes the performance of EMCC against the other clustering methods. On the FRED sequence, our framework delivers the highest overall object discovery accuracy, reaching an $F1$ score of 0.441 and an IoU of 0.647. While ST-Mean Shift yields higher precision, it comes at a computational cost of 517.1 ms per window. In contrast, our method processes the exact same 33 ms temporal segments in just 6.3 ms. This represents a $4.3\times$ improvement over the fastest existing baseline, ST-DBSCAN (27.3 ms), while also delivering better object localization. EMCC bypasses the expensive distance matrix calculations required by traditional density clustering, resulting in major latency and memory improvements, as charted in Figure~\ref{fig:wide_figure_4}. The margin over the baselines follows from the clustering rule rather than from the reduction to one dimension. Morton ordering loses no accuracy here, because bit-boundary splits are re-merged by $d_{\text{link}}$ and false code adjacency is removed by the density check of Eq.~\ref{eq:density}. Our gap threshold re-fits to every window, whereas ST-DBSCAN's $\varepsilon$ and ST-Mean Shift's bandwidth apply a single global scale. ST-HDBSCAN's persistence pruning suppresses sparse objects, which accounts for its recall of 0.281 against 0.517 for EMCC. Qualitative results are presented in Figure~\ref{fig:EMCC_qualitative}.

These architectural advantages generalize effectively to the eTraM dataset. Here, the EMCC framework secures the highest precision (0.378) and matches the top baseline F1 score (0.273) while holding a strict 15.4 ms latency bound, a $2.4\times$ reduction against the next fastest baseline. By consistently outperforming the widely used density-based methods like ST-DBSCAN (36.7 ms) and ST-HDBSCAN (150.1 ms), the framework developed here proves it is robust across different environments. The ability to sustain real-time execution times makes this framework highly practical for real-world perception tasks.

\subsection{Ablation Study}
\label{Ablation}
We present an ablation study on the FRED dataset in Table~\ref{tab:ablation}. We vary each design hyperparameter of SPEF and EMCC independently, holding all others at the baseline ($counter~bits{=}5$, $grid~R{\times}R{=}8$, $\alpha{=}0.8$, $\tau{=}100$ms, $top\text{-}k{=}0.2\%$, smoothing on, $d_{\text{link}}{=}100$px, $bits{=}5$, $dims{=}(x,y,t)$, $p_{\text{gap}}{=}90$, $\alpha_{\text{dens}}{=}12.5$). Baseline values were selected as the configuration achieving the best balance of F1 and latency on the FRED validation set.

\begin{table}[!t]
\centering
\caption{Ablation results for the SPEF and EMCC modules. \textbf{Bold} marks the best value per group. Baseline settings are shaded.}
\label{tab:ablation}
\setlength{\tabcolsep}{3.5pt} 
\renewcommand{\arraystretch}{0.95}

\resizebox{\linewidth}{!}{%
\begin{tabular}[t]{@{}llccccc@{}}
\toprule
\multicolumn{7}{c}{\textbf{SPEF}} \\
\midrule
\textbf{Param} & \textbf{Config} & \textbf{P} & \textbf{R} & \textbf{F1} & \textbf{IoU} & \textbf{Lat.} \\
\midrule
$counter~bits~(b)$ 
     & Raw (No Filter)            & 0.340 & 0.249 & 0.256 & 0.623 & 14.6 \\
     & Probab. (5-bit)            & \textbf{0.427} & 0.488 & 0.434 & 0.643 & \textbf{5.9} \\
\rowcolor{gray!15} & Linear (5-bit)             & 0.416 & 0.\textbf{0.517} & \textbf{0.441} & 0.647 & 6.3 \\
     & Probab. (2-bit)            & 0.416 & 0.343 & 0.353 & \textbf{0.650} & 6.5 \\
     & Linear (2-bit)             & 0.422 & 0.304 & 0.319 & 0.648 & 7.8 \\
\midrule
$\alpha$
     & 0.0 (Spatial)              & \textbf{0.425} & 0.502 & 0.436 & 0.647 & \textbf{6.0} \\
     & 0.5 (Balanced)             & 0.420 & 0.513 & 0.440 & 0.647 & 6.2 \\
\rowcolor{gray!15} & 0.8 (Baseline)             & 0.416 & 0.517 & \textbf{0.441} & 0.647 & 6.3 \\
     & 1.0 (Temporal)             & 0.414 & \textbf{0.520} & \textbf{0.441} & 0.647 & 6.4 \\
\midrule
$\tau$
     & 10ms                       & \textbf{0.427} & 0.513 & \textbf{0.442} & 0.646 & 6.1 \\
     & 33ms                       & 0.420 & 0.516 & 0.441 & \textbf{0.647} & 6.2 \\
     & 50ms                       & 0.418 & 0.516 & 0.441 & \textbf{0.647} & 6.2 \\
\rowcolor{gray!15} & 100ms                      & 0.416 & \textbf{0.517} & 0.441 & \textbf{0.647} & 6.2 \\
\midrule
\rowcolor{gray!15} $top-k$ & 0.2\%                      & \textbf{0.416} & \textbf{0.517} & \textbf{0.441} & 0.647 & \textbf{6.3} \\
     & 0.5\%                      & 0.391 & 0.490 & 0.417 & 0.648 & 6.8 \\
     & 5.0\%                      & 0.363 & 0.415 & 0.373 & \textbf{0.651} & \textbf{6.3} \\
     & 10.0\%                     & 0.363 & 0.404 & 0.369 & \textbf{0.651} & \textbf{6.3} \\
\midrule
$R\times R$
     & 1$\times$1                 & 0.338 & 0.240 & 0.258 & 0.629 & 11.0 \\
     & 2$\times$2                 & 0.380 & 0.364 & 0.351 & \textbf{0.647} & 7.2 \\
     & 4$\times$4                 & 0.415 & 0.471 & 0.422 & \textbf{0.647} & 6.8 \\
    \rowcolor{gray!15} & 8$\times$8                 & \textbf{0.416} & \textbf{0.517} & \textbf{0.441} & \textbf{0.647} & 6.2 \\
    & 12$\times$12                 & 0.406 & 0.499 & 0.428 & 0.646 & \textbf{5.6} \\
\bottomrule
\end{tabular}%
\hspace{15pt}%
\begin{tabular}[t]{@{}llccccc@{}}
\toprule
\multicolumn{7}{c}{\textbf{EMCC}} \\
\midrule
\textbf{Param} & \textbf{Config} & \textbf{P} & \textbf{R} & \textbf{F1} & \textbf{IoU} & \textbf{Lat.} \\
\midrule
$d_{\text{link}}$ & 20px & 0.384 & 0.477 & 0.405 & 0.633 & \textbf{6.2} \\
     & 40px & 0.413 & 0.517 & 0.438 & 0.646 & \textbf{6.2} \\
     & 60px & \textbf{0.416} & \textbf{0.520} & \textbf{0.442} & \textbf{0.647} & \textbf{6.2} \\
     & 80px & \textbf{0.416} & \textbf{0.520} & \textbf{0.442} & \textbf{0.647} & \textbf{6.2} \\
\rowcolor{gray!15} & 100px & \textbf{0.416} & 0.517 & 0.441 & \textbf{0.647} & 6.3 \\
\midrule
\rowcolor{gray!15} $n$ & 5    & \textbf{0.416} & \textbf{0.517} & \textbf{0.441} & \textbf{0.647} & 6.3 \\
     & 10   & 0.415 & 0.505 & 0.435 & 0.645 & 12.7 \\
     & 20   & 0.333 & 0.190 & 0.229 & 0.611 & \textbf{4.5} \\
\midrule
$dims$ & $(x,y)$ & 0.303 & \textbf{0.622} & 0.373 & \textbf{0.650} & \textbf{1.8} \\
\rowcolor{gray!15} & $(x,y,t)$ & \textbf{0.416} & 0.517 & \textbf{0.441} & 0.647 & 6.2 \\
\midrule
$p_{\text{gap}}$  & 70\% & \textbf{0.457} & 0.429 & 0.403 & 0.641 & 6.2 \\
     & 80\% & 0.434 & 0.473 & 0.421 & 0.643 & 6.3 \\
 \rowcolor{gray!15}    & 90\% & 0.416 & \textbf{0.517} & \textbf{0.440} & 0.647 & 6.3 \\
     & 95\% & 0.397 & 0.508 & 0.428 & 0.649 & 5.3 \\
     & 99\% & 0.364 & 0.406 & 0.370 & \textbf{0.654} & \textbf{2.5} \\
\midrule
$\alpha_d$ & 5.0 & 0.397 & \textbf{0.526} & 0.428 & 0.649 & 8.3 \\
\rowcolor{gray!15} & 12.5 & 0.417 & 0.517 & \textbf{0.441} & 0.647 & 6.1 \\
     & 20.0 & \textbf{0.423} & 0.499 & 0.439 & 0.645 & \textbf{5.3} \\
\bottomrule
\end{tabular}%
} 
\end{table}

\noindent\textbf{Without filtering} (EMCC alone), F1 falls to 0.256 at 14.6ms, proving that noise disrupts the density EMCC needs. 

\noindent\textbf{SPEF.} The per-region \textbf{counter} regulates spatial acceptance ($p_o$). A linear 5-bit counter yields the highest F1 (0.441) and recall (0.517). However, a probabilistic 5-bit counter achieves comparable F1 (0.434) at lower latency (5.9ms vs 6.3ms) by accepting state updates stochastically rather than on every event. Under strict memory limits (2-bit depth), the linear counter saturates and fails, dropping F1 to 0.319. The probabilistic counter avoids this by requiring exponentially more events per increment, retaining regional sensitivity and outperforming the linear mode in F1 (0.353 vs 0.319), making it preferable for memory-constrained embedded hardware. The \textbf{temporal memory} ($\alpha$) blends historical correlation with new timings. Performance is stable across $\alpha\in[0.5,1.0]$, with F1 varying by less than 0.001. Dropping history entirely ($\alpha=0.0$) lowers F1 to 0.436, confirming that temporal context contributes to coherence scoring. The \textbf{decay constant} ($\tau$) sets the correlation timescale. Performance is stable across all tested values, with F1 peaking at 0.442 at 10ms and remaining at 0.441 up to 100ms, A larger $\tau$ is preferred as it captures a wider range of edge velocities without penalizing latency, making it more robust across diverse scene dynamics. A $3{\times}3$ \textbf{smoothing} filter has minimal metric impact on this dataset, but it prevents the structural loss of peripheral boundarsy events. Hence, we retain it. The \textbf{top-$k$} threshold is highly sensitive. Raising it from 0.2\% to 10\% drops F1 from 0.441 to 0.369 by discarding moderate-activity regions too early, confirming the spatial gate must remain permissive. Finally, \textbf{grid resolution} ($R$) dictates cell size. Per-pixel ($1{\times}1$) tracking fragments moving edges, dropping F1 to 0.258. An $8{\times}8$ grid maximises F1 (0.441) at 6.2ms, balancing regional event aggregation with boundary preservation. Coarser aggregation at $12{\times}12$ begins to merge distinct object regions, reducing F1 to 0.428.

\vspace{1em}

\noindent\textbf{EMCC.} The \textbf{link distance} ($d_{\text{link}}$) merges adjacent Z-order sub-clusters. F1 plateaus between 60 and 80px at 0.442 but drops to 0.405 at 20px where fragments remain unmerged. Since larger link distances incur negligible latency penalty, the baseline of 100px is preferred as it ensures complete cluster merging across a wider spatial extent, yielding an F1 of 0.441 within 0.001 of the plateau and confirming it is a robust operating point. The \textbf{Morton bit-depth} ($n$) sets coordinate quantization. Counterintuitively, increasing to 20 bits drops F1 to 0.229. With 20 bits allocated to each of the three dimensions $(x,y,t)$, the resulting Morton codes become 60-bit integers. At this scale, the numerical gaps between consecutive events span up to $2^{60}$ values. This massive dynamic range obscures the measurable difference between small intra-cluster gaps and large inter-cluster boundaries, causing percentile-based thresholding to fail. At 5 bits, the gap distribution demonstrates clear bimodality. While this specific optimal depth inherently depends on event density and spatial scale, it highlights a broader architectural principle: EMCC relies on locality-preserving relative ordering rather than absolute coordinate precision. For \textbf{coordinate dimensions}, encoding only space $(x,y)$ merges temporally distinct events into oversized clusters, dropping F1 to 0.373 despite high recall (0.622). Including time $(x,y,t)$ is essential to resolve temporally overlapping objects. The \textbf{gap percentile} ($p_{\text{gap}}$) defines cluster boundaries. Lower values over-fragment clusters, reducing F1 to 0.403 at 70\%, while higher values under-segment, dropping F1 to 0.370 at 99\%. The baseline of 90\% optimally balances these failure modes at F1 of 0.440. Finally, the \textbf{density multiplier} ($\alpha_d$) sets the cluster retention threshold. A loose threshold (5.0) admits noise fragments, inflating latency to 8.3ms. A strict threshold (20.0) reduces latency to 5.3ms but suppresses low-density objects, hurting recall. The baseline of 12.5 optimally balances noise rejection and object retention at F1 of 0.441. Across the tested ranges, F1 varies by at most 0.005 for $\alpha$ and 0.001 for $\tau$, and by less than 0.001 for $d_{\text{link}}$ above 60px. Only the density-scale parameters, top-$k$, counter bits $b$, $n$, and $\alpha_d$, require attention when moving to a new scene.

\section{Discussion and Conclusion}

This work presents a training-free framework for object discovery in event data, combining the probabilistic noise suppression of SPEF with the Morton-ordered clustering of EMCC to operate in real time under strict latency and resource constraints. The central finding is that geometry-based processing grounded in the native spatio-temporal structure of event data is not merely a lightweight alternative but a more accurate and faster solution than existing classical approaches across multiple benchmarks.

SPEF ranks first or second across six of eight noise conditions on the E-MLB dataset benchmark, without any offline training, narrowing the performance gap with learned denoisers substantially.  On the FRED dataset, EMCC delivers a $23.2\%$ improvement in F1 score and an $8.9\%$ increase in IoU over the ST-DBSCAN, which is also the next fastest baseline, at a $4.3\times$ latency reduction. The framework generalizes to the eTraM dataset, matching top baseline accuracy while recording the lowest execution time across all evaluated methods. Across two benchmarks, Morton-ordered gap detection does not sacrifice accuracy and in fact advances it, while reducing computational cost to a degree that makes deployment on resource-constrained edge hardware practical. Furthermore, because the gap-based clustering follows from explicit geometric rules, every decision in the pipeline is traceable to a real spatio-temporal structure in the event stream, which makes failure diagnosis straightforward. Discovery quality is also sensitive to the signal-to-noise ratio of the input stream, and object identity is not maintained across instances. These are well-understood constraints of annotation-free discovery and define clear directions for future work. Field deployment on embedded platforms under real operating conditions will establish the practical boundary of the pipeline, while lightweight online tracking would extend it toward a complete, training-free perception stack for real-time edge deployment.

\section*{Acknowledgements}

The authors gratefully acknowledge Emeritus Professor Mandyam Srinivasan for his generous guidance throughout this work, including detailed manuscript review and participation in research discussions.


\clearpage
\setcounter{section}{0}
\setcounter{figure}{0}
\setcounter{table}{0}
\setcounter{equation}{0}
\renewcommand{\thesection}{S\arabic{section}}
\renewcommand{\thefigure}{S\arabic{figure}}
\renewcommand{\thetable}{S\arabic{table}}
\renewcommand{\theequation}{S\arabic{equation}}

\begin{center}
{\LARGE \sc Supplementary Material}
\end{center}
\vspace{1em}

\section{Pipeline Modularity and Compatibility with Ego-Motion Compensation}
SPEF and EMCC are designed as downstream modules that operate on pre-processed event streams, deliberately decoupling coordinate stabilisation from object discovery. This serves two purposes. First, it enables stage-isolated evaluation: since ego-motion compensation methods vary substantially in accuracy and hardware dependency, coupling these stages would conflate their error sources and obscure the contribution of SPEF and EMCC individually. Second, the modular interface makes the framework compatible with the full spectrum of upstream compensation strategies, including IMU-based rotational warping~\cite{zhao, Survey}, contrast maximisation~\cite{Cmax1, Cmax2}, and learning-based motion compensation~\cite{zhu}, without requiring any modification to either module.

This separation also yields a structural advantage for EMCC specifically. When ego-motion compensation is applied upstream, events from static structure share spatial coordinates across time, widening the Morton-space gap between background clutter and independently moving targets. The adaptive threshold $\tau = \text{percentile}(\Delta, p_{\text{gap}})$ therefore becomes more discriminative, as the gap distribution grows bimodal, consistent with the theoretical properties of the Z-order curve under spatially coherent inputs~\cite{z-order1}. This provides a principled account of why pre-compensated input improves clustering performance without making ego-motion compensation a hard dependency of the method.

\section{EMCC: Full Implementation Details}
\label{SUPP: EMCC Algorithm}

Algorithm~\ref{alg:EMCC} in the main paper presents a compact summary of EMCC. 
Algorithm~\ref{alg:EMCC_Full} here expands this into a fully specified procedure, 
making explicit the parameter inputs, the density filter condition, and the aggregation 
and NMS steps that are referenced but not detailed in the main text. Additional qualitative comparison is provided in Figure \ref{fig:supp_quali}.

\setcounter{algorithm}{0}  
\renewcommand{\thealgorithm}{SA\arabic{algorithm}}
\begin{algorithm}[!t]
\caption{EMCC: Full Procedure}
\label{alg:EMCC_Full}
\begin{algorithmic}[1]
\Require Events $\mathcal{E} = \{(x_i, y_i, t_i)\}_{i=1}^{N}$, resolution $W \times H$,
bit depth $n$, gap percentile $p_{\text{gap}}$, box percentiles $p_{\text{low}},
p_{\text{high}}$, density multiplier $\alpha$, merge distance $d_{\text{link}}$
\Ensure Bounding boxes $\mathcal{B}$

\State $d_{\text{global}} \gets N / (W \times H)$

\vspace{4pt}
\Comment{Normalise to a common $n$-bit integer range before encoding}
\For{each $(x, y, t) \in \mathcal{E}$}
    \State $x' \gets \lfloor (x / W) \cdot (2^n - 1) \rfloor$
    \State $y' \gets \lfloor (y / H) \cdot (2^n - 1) \rfloor$
    \State $t' \gets \lfloor ((t - t_{\min}) / (t_{\max} - t_{\min})) \cdot (2^n - 1) \rfloor$
\EndFor

\vspace{4pt}
\Comment{Encode and sort: nearby events in 3D space map to nearby scalars}
\For{each normalised $(x', y', t')$}
    \State $m_i \gets \text{MortonEncode}(x', y', t')$
\EndFor
\State $\sigma \gets \text{argsort}(\mathbf{m})$

\vspace{4pt}
\Comment{Large gaps mark object boundaries; threshold adapts to stream density}
\State $\Delta_i \gets m_{\sigma(i+1)} - m_{\sigma(i)}, \quad i = 1, \ldots, N-1$
\State $\tau \gets \text{percentile}(\Delta,\, p_{\text{gap}})$
\State Split at positions where $\Delta_i > \tau$ to obtain $\{\mathcal{C}_1,
\mathcal{C}_2, \ldots\}$

\vspace{4pt}
\Comment{Discard clusters whose event density falls below the global scene rate}
\State $\mathcal{B}_{\text{raw}} \gets \emptyset$, \quad $\mathcal{S} \gets \emptyset$
\For{each cluster $\mathcal{C}_k$}
    \State $(x_{\min}, y_{\min}, x_{\max}, y_{\max}) \gets$ percentile bounds at
    $p_{\text{low}},\, p_{\text{high}}$
    \State $A \gets (x_{\max} - x_{\min})(y_{\max} - y_{\min})$
    \If{$|\mathcal{C}_k| / A \geq \alpha \cdot d_{\text{global}}$}
        \State $\mathcal{B}_{\text{raw}} \gets \mathcal{B}_{\text{raw}} \cup
        \{(x_{\min}, y_{\min}, x_{\max}, y_{\max})\}$
        \State $\mathcal{S} \gets \mathcal{S} \cup \{|\mathcal{C}_k|\}$
    \EndIf
\EndFor

\vspace{4pt}
\Comment{Merge sub-clusters split by Z-order bit boundaries, scored by cluster size}
\State $\mathcal{B}_{\text{agg}} \gets \text{SpatialAggregation}(\mathcal{B}_{\text{raw}}
,\, d_{\text{link}})$
\State $\mathcal{B} \gets \text{NMS}(\mathcal{B}_{\text{agg}},\, \mathcal{S})$
\State \Return $\mathcal{B}$
\end{algorithmic}
\end{algorithm}

\section{Hyperparameter Ranges}
\label{SUPP: Parameter Tuning}

Table~\ref{tab:parameter and ranges} defines the search spaces for all method-specific and shared hyperparameters, explored using Optuna~\cite{Optuna} with 1000 trials per algorithm per dataset. The trial budget was determined empirically as the point beyond which the TPE sampler showed no meaningful improvement in composite score across held-out validation runs. To prevent overfitting, hyperparameter search is conducted on a held-out validation split disjoint from the test sequences used for performance reporting in the main text. Search bounds are tailored to the mechanics of each method and the characteristics of each dataset.

ST-DBSCAN~\cite{ST-DBSCAN} uses a global fixed radius, which struggles with the scale variation introduced by object distance, speed, and lighting. We tune $eps \in [0.005, 0.3]$ to cover both tight and diffuse clusters, with the upper bound extended for eTraM to accommodate wider spatial variance in traffic scenes. ST-HDBSCAN~\cite{mcinnes2017hdbscan, HDBSCAN} constructs a hierarchical dendrogram to detect variable density levels, but empirically the hierarchy adds little benefit above 200 events and fragments isolated detections below 5, so we restrict $min\_cluster\_size \in [5, 200]$ accordingly. For ST-MeanShift~\cite{meanshift}, the bandwidth controls the smoothing window and is the single most sensitive parameter: values that are too small fragment single objects while values that are too large merge adjacent targets. We tune $bandwidth \in [0.02, 0.5]$ with dataset-specific bounds reflecting the difference in object density between FRED and eTraM.

For EMCC, events are sorted by 3D Morton code such that consecutive events from the same object produce small gaps while transitions between distinct spatial clusters produce large ones. The partition threshold is set at $cluster\_gap\_percentile \in [60.0, 99.8]$, adapting to scene density rather than requiring a globally fixed value. The lower bound for eTraM is relaxed relative to FRED to account for higher event density and closer object proximity in traffic scenes.

After clustering, a shared set of geometric constraints filters the resulting bounding boxes without learned semantic features. Although shared geometric parameters are optimised jointly with method-specific ones, each baseline receives its own independently optimised shared parameter set, ensuring no configuration is disadvantaged by a suboptimal box filtering stage. $merge\_margin \in [5.0, 100.0]$ combines fragmented boxes from the same object across the scale range of both datasets. $box\_percentiles$ tighten boundaries to dense event cores, and $box\_margin \in [0.0, 30.0]$ expands them slightly to ensure complete object coverage.

\setcounter{table}{0}  
\renewcommand{\thetable}{ST\arabic{table}}
\begin{table}[!t]
\centering
\caption{Hyperparameter search ranges for the evaluated clustering algorithms on FRED~\cite{magrini2025fred} and eTraM~\cite{etram}. Shared hyperparameters govern bounding box generation and are jointly optimised alongside method-specific parameters, with each algorithm receiving its own independently optimised shared parameter set.}
\label{tab:parameter and ranges}
\begin{tabularx}{\columnwidth}{
    >{\raggedright\arraybackslash}p{0.4\columnwidth}
    >{\raggedright\arraybackslash}X
    >{\raggedright\arraybackslash}X
}
\toprule
\textbf{Hyperparameters} & \textbf{FRED~\cite{magrini2025fred}} & \textbf{eTraM~\cite{etram}} \\
\midrule
\textbf{ST-DBSCAN} & & \\
eps & 0.005--0.1 & 0.005--0.3 \\
min samples & 3--50 & 3--50 \\
\midrule
\textbf{ST-HDBSCAN} & & \\
min cluster size & 5--60  & 5--200  \\
min samples & 1--30 & 1--30 \\
\midrule
\textbf{ST-MeanShift} & & \\
bandwidth & 0.02--0.2  & 0.05--0.5  \\
\midrule
\textbf{EMCC (Ours)} & & \\
cluster gap percentile & 80.0--99.8 & 60.0--99.8 \\
density multiplier & 0.1--15 & 0.05--10 \\
\midrule
\textbf{Shared Parameters} & & \\
merge margin & 10--100 & 5--75 \\
box percentile low & 0--5 & 1--20 \\
box percentile high & 95--100 & 90--99 \\
box margin & 0--20 & 1--30 \\
\bottomrule
\end{tabularx}
\end{table}

\section{Hyperparameter Tuning --- Optimization Framework}
\label{Sec: Parametrt Tuning}

Hyperparameters are optimised using Optuna with the Tree-structured Parzen Estimator (TPE) sampler~\cite{Optuna, TPE}, as detailed in Algorithm~\ref{alg:hyperopt}. The same pipeline is applied to SPEF+EMCC and all baselines to ensure a fair comparison. The optimization objective maximises a composite score that balances spatial accuracy, temporal coverage, and false positive suppression:

\setcounter{equation}{0}  
\renewcommand{\theequation}{SE\arabic{equation}}
\begin{equation}
\text{Composite Score} = (\text{Precision} + \text{Recall} + \text{mIoU} + \lambda \cdot \mathcal{F}_{\text{det}}) \cdot \mathcal{P}_{\text{FP}}
\end{equation}

\textbf{Accuracy metrics.} Precision, Recall, and mIoU are computed at an IoU threshold of 0.5 for true positive matches, following the convention established in event-based object detection benchmarks~\cite{etram, magrini2025fred}. This combination ensures the optimised configuration reaches a balance between accurate localisation and reliable object discovery.

\renewcommand{\thealgorithm}{SA\arabic{algorithm}}
\begin{algorithm}[!b]
\caption{Bayesian Hyperparameter Optimization for Event-based Detection}
\label{alg:hyperopt}
\begin{algorithmic}[1]
\Require Event stream $\mathcal{E}$, ground-truth $\mathcal{G}$, parameter space $\Theta$, trial budget $N_{\text{trials}}$
\Ensure Optimal parameters $\theta^*$ and best score $\mathcal{S}^*$
\State Initialise SPEF filter and clustering module and validate on sample data
\For{$i = 1$ \textbf{to} $N_{\text{trials}}$}
    \State Sample candidate parameters $\theta_i$ from $\Theta$
    \State Apply SPEF filtering to event stream
    \State Cluster filtered events via selected algorithm
    \State Generate and filter bounding boxes
    \State Evaluate predictions against $\mathcal{G}$ to compute TP, FP, FN, and IoUs
    \State Compute base score: $\mathcal{S}_{\text{base}} = \text{Precision} + \text{Recall} + \text{mIoU} + 0.1 \cdot \mathcal{F}_{\text{det}}$
    \State Compute false positive penalty $\mathcal{P}_{\text{FP}}$ from $R_{\text{FP}}$
    \State Compute composite score: $\mathcal{S}_i = \mathcal{S}_{\text{base}} \cdot \mathcal{P}_{\text{FP}}$
    \State Report $\mathcal{S}_i$ to optimizer
\EndFor
\State $\theta^* \gets \arg\max_i \mathcal{S}_i$
\State \Return $\theta^*, \mathcal{S}^*$
\end{algorithmic}
\end{algorithm}

\textbf{Detection frequency ($\mathcal{F}_{\text{det}}$).}
\begin{equation}
\mathcal{F}_{\text{det}} = \frac{\text{Frames with valid predictions}}{\text{Total frames with ground-truth objects}}
\end{equation}
This measures temporal consistency across the sequence. The weight $\lambda = 0.1$ is deliberately small so that $\mathcal{F}_{\text{det}}$ acts as a tie-breaker among configurations with similar spatial accuracy, rather than allowing the optimizer to inflate bounding boxes to sustain detection rate at the cost of localisation precision. Sensitivity to $\lambda$ was verified by sweeping values in $[0.01, 0.5]$, confirming that the relative ranking of configurations remains stable across this range.

\textbf{False positive penalty ($\mathcal{P}_{\text{FP}}$).} Clustering methods do not perform semantic feature extraction, making them susceptible to detecting unannotated background motions such as moving vegetation. To suppress this without bottlenecking recall, we apply a bounded multiplicative penalty based on the false positive ratio $R_{\text{FP}} = \text{FP} / (\text{TP} + 1)$:
\begin{equation}
\mathcal{P}_{\text{FP}} = 
\begin{cases} 
1.0, & \text{if } R_{\text{FP}} \le 1.3 \\
\frac{1.3}{R_{\text{FP}}}, & \text{otherwise}
\end{cases}
\end{equation}
The tolerance threshold of 1.3 was determined by inspecting the false positive distribution across both datasets. In practice, a ratio of up to 1.3 false positives per true positive reflects the baseline level of background clutter that non-semantic clustering methods produce under normal scene conditions, such as sparse moving foliage or sensor noise bursts. Beyond this point, the false positive count grows disproportionately relative to true detections, indicating systematic over-detection rather than unavoidable background response. The multiplicative form of $\mathcal{P}_{\text{FP}}$ ensures that once this threshold is crossed, the score penalty scales directly with the degree of over-detection, preventing the optimizer from trading localisation quality for an inflated detection count. A strict zero-tolerance constraint was avoided because it would force the optimizer toward overly conservative parameters that suppress valid detections during moments of sparse object activity. Algorithm~\ref{alg:hyperopt} details the complete optimization workflow.

\section{Probabilistic vs. Deterministic Event Acceptance: An Empirical Analysis}

As introduced in Section~\ref{SPEF_Method} of the main paper, SPEF accepts events probabilistically: an event is accepted if $r \sim \mathcal{U}(0,1)$ satisfies $r < p_o \cdot p_t$, where $p_o$ and $p_t$ are spatial and temporal acceptance probabilities computed per-event from local scene conditions. Unlike fixed thresholds, which apply a uniform acceptance boundary regardless of scene state, this probabilistic formulation adapts the decision boundary to local spatio-temporal activity. Comprehensive evaluation on the FRED dataset confirms a consistent advantage of probabilistic acceptance over fixed thresholds across diverse scene conditions. We examine three representative cases that illuminate the mechanisms driving this advantage.

\textbf{Instance 63: multiple objects.} During the first 500 frames both objects are inactive and $p_o \cdot p_t$ remains near-zero, suppressing false positives from background noise. When objects enter the scene, acceptance probability rises sharply and tracks their coherent motion, dropping again during quiescent periods. As shown in Figure~\ref{fig:pkeep}(a)(ii), RNG achieves performance comparable to the best fixed threshold ($\theta_{\text{fix}} = 0.4$) but without the systematic degradation observed at higher thresholds. Fixed thresholds above 0.4 exhibit monotonic performance loss, exposing the brittleness of uniform acceptance boundaries under varying object activity.

\setcounter{figure}{0}  
\renewcommand{\thefigure}{SF\arabic{figure}}
\begin{figure*}[!tpbh]
  \centering
  \includegraphics[width=\textwidth]{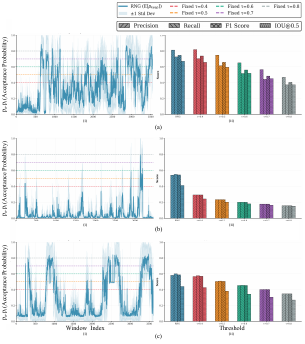}
  \caption{Comparison of probabilistic (RNG) and fixed-threshold acceptance across three representative scenarios: (a) multiple objects (Instance 63), (b) distant sparse target (Instance 90), and (c) dynamic proximity changes (Instance 154). Subplots (i) show acceptance probability over time and subplots (ii) report Precision, Recall, F1, and IoU@50. RNG (blue) adapts dynamically to scene activity, outperforming fixed thresholds at $\theta_{\text{fix}} \in \{0.4, 0.5, 0.6, 0.7, 0.8\}$ across all scenarios.}
  \label{fig:pkeep}
\end{figure*}

\textbf{Instance 90: sparse distant target.} The target produces weak spatial and temporal coherence throughout, keeping acceptance probability consistently low. This is the appropriate response to a genuinely sparse signal rather than an artifact of aggressive filtering. RNG significantly outperforms all fixed thresholds ($\theta_{\text{fix}} \in \{0.4, \ldots, 0.8\}$), which over-reject events and degrade both precision and recall. This instance represents the largest observed performance margin and demonstrates that probabilistic modulation is most critical precisely when signal coherence is weakest.

\textbf{Instance 154: dynamic approach and recession.} The object alternates between approaching the camera, driving acceptance probability to peak, and receding, causing it to drop sharply. RNG tracks this oscillation window-by-window, maintaining high F1 and IoU@50 throughout. Fixed thresholds cannot respond to this rate of change and exhibit precision-recall trade-offs that worsen as the object recedes.

\setcounter{figure}{1}  
\renewcommand{\thefigure}{SF\arabic{figure}}
\begin{figure*}[!hptb]
  \centering
  \includegraphics[width=\textwidth]{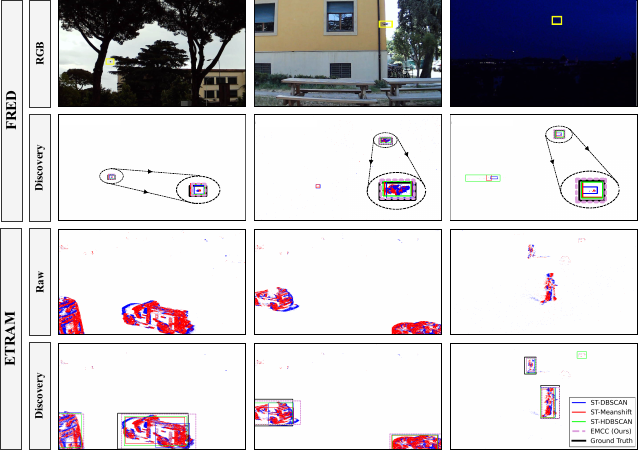}
  \caption{\textbf{Qualitative comparison of object detection outputs on the FRED and eTraM datasets:}
additional samples. Yellow boxes on RGB frames of FRED provide visual reference, with black-dashed insets showing zoomed-in views of discovered objects. Bounding box proposals are displayed for ST-DBSCAN (blue), ST-MeanShift (red), ST-HDBSCAN (green), EMCC (purple-dashed), and Ground Truth (black). These supplementary examples extend the qualitative analysis presented in the main paper, demonstrating EMCC's performance consistency across diverse scene configurations and dynamic activity levels.}
  \label{fig:supp_quali}
\end{figure*}

Across all three cases, covering multiple objects, sparse signals, and dynamic motion, RNG consistently maintains higher F1 scores while the best fixed threshold ($\theta_{\text{fix}} = 0.4$) degrades under conditions it was not tuned for. The acceptance probability traces in Figure~\ref{fig:pkeep}(i) confirm that this is genuine adaptability: SPEF adjusts its acceptance rate to match local spatio-temporal coherence without any manual intervention. Soft probabilistic boundaries remove the brittleness of hard thresholding and allow the filter to generalise across scene conditions without retuning.

\newpage
\bibliographystyle{unsrt}
\bibliography{egbib}

@ARTICLE{Survey,
  author={Gallego, Guillermo and Delbrück, Tobi and Orchard, Garrick and Bartolozzi, Chiara and Taba, Brian and Censi, Andrea and Leutenegger, Stefan and Davison, Andrew J. and Conradt, Jörg and Daniilidis, Kostas and Scaramuzza, Davide},
  journal={IEEE Transactions on Pattern Analysis and Machine Intelligence}, 
  title={Event-Based Vision: A Survey}, 
  year={2022},
  volume={44},
  number={1},
  pages={154-180},
  doi={10.1109/TPAMI.2020.3008413}}

@InProceedings{ASCNN,
author="Messikommer, Nico
and Gehrig, Daniel
and Loquercio, Antonio
and Scaramuzza, Davide",
editor="Vedaldi, Andrea
and Bischof, Horst
and Brox, Thomas
and Frahm, Jan-Michael",
title="Event-Based Asynchronous Sparse Convolutional Networks",
booktitle="Computer Vision -- ECCV 2020",
year="2020",
publisher="Springer International Publishing",
address="Cham",
pages="415--431",
isbn="978-3-030-58598-3"
}

@ARTICLE{Evcam,
  author={Lichtsteiner, Patrick and Posch, Christoph and Delbruck, Tobi},
  journal={IEEE Journal of Solid-State Circuits}, 
  title={A 128$\times$ 128 120 dB 15 $\mu$s Latency Asynchronous Temporal Contrast Vision Sensor}, 
  year={2008},
  volume={43},
  number={2},
  pages={566-576},
  doi={10.1109/JSSC.2007.914337}}

@INPROCEEDINGS{moving_obj,
  author={Mitrokhin, Anton and Fermüller, Cornelia and Parameshwara, Chethan and Aloimonos, Yiannis},
  booktitle={2018 IEEE/RSJ International Conference on Intelligent Robots and Systems (IROS)}, 
  title={Event-Based Moving Object Detection and Tracking}, 
  year={2018},
  volume={},
  number={},
  pages={1-9},
  doi={10.1109/IROS.2018.8593805}}

@ARTICLE{blob,
  author={Wang, Ziwei and Molloy, Timothy and van Goor, Pieter and Mahony, Robert},
  journal={IEEE Transactions on Robotics}, 
  title={Asynchronous Blob Tracker for Event Cameras}, 
  year={2024},
  volume={40},
  number={},
  pages={4750-4767},
  doi={10.1109/TRO.2024.3454410}}

@ARTICLE{DAVIS,
  author={Brandli, Christian and Berner, Raphael and Yang, Minhao and Liu, Shih-Chii and Delbruck, Tobi},
  journal={IEEE Journal of Solid-State Circuits}, 
  title={A 240 × 180 130 dB 3 µs Latency Global Shutter Spatiotemporal Vision Sensor}, 
  year={2014},
  volume={49},
  number={10},
  pages={2333-2341},
  doi={10.1109/JSSC.2014.2342715}}

@article{discovery1,
  title={Unsupervised object discovery: A comparison},
  author={Tuytelaars, Tinne and Lampert, Christoph H and Blaschko, Matthew B and Buntine, Wray},
  journal={International journal of computer vision},
  volume={88},
  number={2},
  pages={284--302},
  year={2010},
  publisher={Springer}
}

@inproceedings{discovery2,
 author = {Vo, Van Huy and Sizikova, Elena and Schmid, Cordelia and P\'{e}rez, Patrick and Ponce, Jean},
 booktitle = {Advances in Neural Information Processing Systems},
 editor = {M. Ranzato and A. Beygelzimer and Y. Dauphin and P.S. Liang and J. Wortman Vaughan},
 pages = {16764--16778},
 publisher = {Curran Associates, Inc.},
 title = {Large-Scale Unsupervised Object Discovery},
 url = {https://proceedings.neurips.cc/paper_files/paper/2021/file/8bf1211fd4b7b94528899de0a43b9fb3-Paper.pdf},
 volume = {34},
 year = {2021}
}

@InProceedings{Discovery3,
author = {Kwak, Suha and Cho, Minsu and Laptev, Ivan and Ponce, Jean and Schmid, Cordelia},
title = {Unsupervised Object Discovery and Tracking in Video Collections},
booktitle = {Proceedings of the IEEE International Conference on Computer Vision (ICCV)},
month = {December},
year = {2015}
}

@article{Discovery4,
  title={Unsupervised object localization in the era of self-supervised vits: A survey},
  author={Sim{\'e}oni, Oriane and Zablocki, {\'E}loi and Gidaris, Spyros and Puy, Gilles and P{\'e}rez, Patrick},
  journal={International Journal of Computer Vision},
  volume={133},
  number={2},
  pages={781--808},
  year={2025},
  publisher={Springer}
}

@InProceedings{moving2,
    author    = {Mondal, Anindya and R, Shashant and Giraldo, Jhony H. and Bouwmans, Thierry and Chowdhury, Ananda S.},
    title     = {Moving Object Detection for Event-Based Vision Using Graph Spectral Clustering},
    booktitle = {Proceedings of the IEEE/CVF International Conference on Computer Vision (ICCV) Workshops},
    month     = {October},
    year      = {2021},
    pages     = {876-884}
}

@INPROCEEDINGS{moving3,
  author={Mondal, Anindya and Das, Mayukhmali},
  booktitle={2021 IEEE 8th Uttar Pradesh Section International Conference on Electrical, Electronics and Computer Engineering (UPCON)}, 
  title={Moving Object Detection for Event-based Vision using k-means Clustering}, 
  year={2021},
  volume={},
  number={},
  pages={1-6},
  doi={10.1109/UPCON52273.2021.9667636}}

@INPROCEEDINGS{moving4,
  author={Shu, Yuanjun and Sui, Yunfeng and Zhao, Shixuan and Cheng, Zhi and Liu, Weiqian},
  booktitle={2021 7th International Conference on Computer and Communications (ICCC)}, 
  title={Small Moving Object Detection and Tracking Based on Event Signals}, 
  year={2021},
  volume={},
  number={},
  pages={792-796},
  doi={10.1109/ICCC54389.2021.9674247}}

@Article{moving5,
AUTHOR = {Zhao, Jiang and Ji, Shilong and Cai, Zhihao and Zeng, Yiwen and Wang, Yingxun},
TITLE = {Moving Object Detection and Tracking by Event Frame from Neuromorphic Vision Sensors},
JOURNAL = {Biomimetics},
VOLUME = {7},
YEAR = {2022},
NUMBER = {1},
ARTICLE-NUMBER = {31},
URL = {https://www.mdpi.com/2313-7673/7/1/31},
PubMedID = {35323188},
ISSN = {2313-7673},
DOI = {10.3390/biomimetics7010031}
}

@INPROCEEDINGS{prereq1,
  author={Lu, Xiuyuan and Zhou, Yi and Shen, Shaojie},
  booktitle={2021 IEEE/RSJ International Conference on Intelligent Robots and Systems (IROS)}, 
  title={Event-based Motion Segmentation by Cascaded Two-Level Multi-Model Fitting}, 
  year={2021},
  volume={},
  number={},
  pages={4445-4452},
  doi={10.1109/IROS51168.2021.9636307}}

@InProceedings{prereq2,
author = {Stoffregen, Timo and Gallego, Guillermo and Drummond, Tom and Kleeman, Lindsay and Scaramuzza, Davide},
title = {Event-Based Motion Segmentation by Motion Compensation},
booktitle = {Proceedings of the IEEE/CVF International Conference on Computer Vision (ICCV)},
month = {October},
year = {2019}
}

@INPROCEEDINGS{prereq3,
  author={Mitrokhin, Anton and Ye, Chengxi and Fermüller, Cornelia and Aloimonos, Yiannis and Delbruck, Tobi},
  booktitle={2019 IEEE/RSJ International Conference on Intelligent Robots and Systems (IROS)}, 
  title={EV-IMO: Motion Segmentation Dataset and Learning Pipeline for Event Cameras}, 
  year={2019},
  volume={},
  number={},
  pages={6105-6112},
  doi={10.1109/IROS40897.2019.8968520}}

@article{GMCM,
  title={A motion denoising algorithm with Gaussian self-adjusting threshold for event camera},
  author={Lin, Wanmin and Li, Yuhui and Xu, Chen and Liu, Lilin},
  journal={The Visual Computer},
  volume={40},
  number={9},
  pages={6567--6580},
  year={2024},
  publisher={Springer}
}

@INPROCEEDINGS{RobMohoney,
  author={Wang, Ziwei and Yuan, Dingran and Ng, Yonhon and Mahony, Robert},
  booktitle={2022 International Conference on Robotics and Automation (ICRA)}, 
  title={A Linear Comb Filter for Event Flicker Removal}, 
  year={2022},
  volume={},
  number={},
  pages={398-404},
  doi={10.1109/ICRA46639.2022.9812003}}

@ARTICLE{E-MLB,
  author={Ding, Saizhe and Chen, Jinze and Wang, Yang and Kang, Yu and Song, Weiguo and Cheng, Jie and Cao, Yang},
  journal={IEEE Transactions on Multimedia}, 
  title={E-MLB: Multilevel Benchmark for Event-Based Camera Denoising}, 
  year={2024},
  volume={26},
  number={},
  pages={65-76},
  doi={10.1109/TMM.2023.3260638}}

@article{noisy,
author = {Meriem Ben Miled  and Wenwen Liu  and Yuanchang Liu },
title = {Adaptive Unsupervised Learning-Based 3D Spatiotemporal Filter for Event-Driven Cameras},
journal = {Research},
volume = {7},
number = {},
pages = {0330},
year = {2024},
doi = {10.34133/research.0330},
URL = {https://spj.science.org/doi/abs/10.34133/research.0330},
eprint = {https://spj.science.org/doi/pdf/10.34133/research.0330}}

@misc{voxel1,
      title={Unsupervised Event-based Learning of Optical Flow, Depth, and Egomotion}, 
      author={Alex Zihao Zhu and Liangzhe Yuan and Kenneth Chaney and Kostas Daniilidis},
      year={2018},
      eprint={1812.08156},
      archivePrefix={arXiv},
      primaryClass={cs.CV},
      url={https://arxiv.org/abs/1812.08156}, 
}

@misc{voxel2,
      title={E-RAFT: Dense Optical Flow from Event Cameras}, 
      author={Mathias Gehrig and Mario Millhäusler and Daniel Gehrig and Davide Scaramuzza},
      year={2021},
      eprint={2108.10552},
      archivePrefix={arXiv},
      primaryClass={cs.CV},
      url={https://arxiv.org/abs/2108.10552}, 
}

@ARTICLE{timesurfaces1,
  author={Lagorce, Xavier and Orchard, Garrick and Galluppi, Francesco and Shi, Bertram E. and Benosman, Ryad B.},
  journal={IEEE Transactions on Pattern Analysis and Machine Intelligence}, 
  title={HOTS: A Hierarchy of Event-Based Time-Surfaces for Pattern Recognition}, 
  year={2017},
  volume={39},
  number={7},
  pages={1346-1359},
  doi={10.1109/TPAMI.2016.2574707}}

@misc{timesurfaces2,
      title={HATS: Histograms of Averaged Time Surfaces for Robust Event-based Object Classification}, 
      author={Amos Sironi and Manuele Brambilla and Nicolas Bourdis and Xavier Lagorce and Ryad Benosman},
      year={2018},
      eprint={1803.07913},
      archivePrefix={arXiv},
      primaryClass={cs.CV},
      url={https://arxiv.org/abs/1803.07913}, 
}

@misc{timesurfaces3,
      title={Speed Invariant Time Surface for Learning to Detect Corner Points with Event-Based Cameras}, 
      author={Jacques Manderscheid and Amos Sironi and Nicolas Bourdis and Davide Migliore and Vincent Lepetit},
      year={2019},
      eprint={1903.11332},
      archivePrefix={arXiv},
      primaryClass={cs.CV},
      url={https://arxiv.org/abs/1903.11332}, 
}

@INPROCEEDINGS{CNN1,
  author={Ghosh, Rohan and Mishra, Abhishek and Orchard, Garrick and Thakor, Nitish V.},
  booktitle={2014 IEEE Biomedical Circuits and Systems Conference (BioCAS) Proceedings}, 
  title={Real-time object recognition and orientation estimation using an event-based camera and CNN}, 
  year={2014},
  volume={},
  number={},
  pages={544-547},
  doi={10.1109/BioCAS.2014.6981783}}

@misc{CNN3,
      title={Deep Learning for Event-based Vision: A Comprehensive Survey and Benchmarks}, 
      author={Xu Zheng and Yexin Liu and Yunfan Lu and Tongyan Hua and Tianbo Pan and Weiming Zhang and Dacheng Tao and Lin Wang},
      year={2024},
      eprint={2302.08890},
      archivePrefix={arXiv},
      primaryClass={cs.CV},
      url={https://arxiv.org/abs/2302.08890}, 
}

@inproceedings{AEGNN,
  title={Aegnn: Asynchronous event-based graph neural networks},
  author={Schaefer, Simon and Gehrig, Daniel and Scaramuzza, Davide},
  booktitle={Proceedings of the IEEE/CVF conference on computer vision and pattern recognition},
  pages={12371--12381},
  year={2022}
}

@ARTICLE{EVGNN,
  author={Yang, Yufeng and Kneip, Adrian and Frenkel, Charlotte},
  journal={IEEE Transactions on Circuits and Systems for Artificial Intelligence}, 
  title={EvGNN: An Event-Driven Graph Neural Network Accelerator for Edge Vision}, 
  year={2025},
  volume={2},
  number={1},
  pages={37-50},
  doi={10.1109/TCASAI.2024.3520905}}

@article{RNN1,
  title={Phased lstm: Accelerating recurrent network training for long or event-based sequences},
  author={Neil, Daniel and Pfeiffer, Michael and Liu, Shih-Chii},
  journal={Advances in neural information processing systems},
  volume={29},
  year={2016}
}

@InProceedings{RNN2,
author="Cannici, Marco
and Ciccone, Marco
and Romanoni, Andrea
and Matteucci, Matteo",
editor="Vedaldi, Andrea
and Bischof, Horst
and Brox, Thomas
and Frahm, Jan-Michael",
title="A Differentiable Recurrent Surface for Asynchronous Event-Based Data",
booktitle="Computer Vision -- ECCV 2020",
year="2020",
publisher="Springer International Publishing",
address="Cham",
pages="136--152",
isbn="978-3-030-58565-5"
}

@article{GNNAutomotive,
  title={Low-latency automotive vision with event cameras},
  author={Gehrig, Daniel and Scaramuzza, Davide},
  journal={Nature},
  volume={629},
  number={8014},
  pages={1034--1040},
  year={2024},
  publisher={Nature Publishing Group UK London}
}

@inproceedings{SNN1,
  title={Spiking-{YOLO}: Spiking Neural Network for Energy-Efficient Object Detection},
  author={Kim, Seijoon and Park, Seongsik and Na, Byunggook and Yoon, Sungroh},
  booktitle={Proceedings of the AAAI Conference on Artificial Intelligence},
  volume={34},
  pages={11270--11277},
  year={2020},
  doi={10.1609/aaai.v34i07.6787}
}

@inproceedings{SNN2,
  title={Object detection with spiking neural networks on automotive event data},
  author={Cordone, Lo{\"\i}c and Miramond, Beno{\^\i}t and Thierion, Philippe},
  booktitle={2022 International Joint Conference on Neural Networks (IJCNN)},
  pages={1--8},
  year={2022},
  organization={IEEE}
}

@inproceedings{SNN3,
  title={Spikepoint: An efficient point-based spiking neural network for event cameras action recognition},
  author={Ren, Hongwei and Zhou, Yue and Lin, Xiaopeng and Huang, Yulong and Fu, Haotian and Song, Jie and Cheng, Bojun},
  booktitle={International Conference on Learning Representations},
  volume={2024},
  pages={27827--27846},
  year={2024}
}

@article{RED,
  title={Learning to detect objects with a 1 megapixel event camera},
  author={Perot, Etienne and De Tournemire, Pierre and Nitti, Davide and Masci, Jonathan and Sironi, Amos},
  journal={Advances in Neural Information Processing Systems},
  volume={33},
  pages={16639--16652},
  year={2020}
}

@InProceedings{RVT,
    author    = {Gehrig, Mathias and Scaramuzza, Davide},
    title     = {Recurrent Vision Transformers for Object Detection With Event Cameras},
    booktitle = {Proceedings of the IEEE/CVF Conference on Computer Vision and Pattern Recognition (CVPR)},
    month     = {June},
    year      = {2023},
    pages     = {13884-13893}
}

@misc{Gen1,
      title={A Large Scale Event-based Detection Dataset for Automotive}, 
      author={Pierre de Tournemire and Davide Nitti and Etienne Perot and Davide Migliore and Amos Sironi},
      year={2020},
      eprint={2001.08499},
      archivePrefix={arXiv},
      primaryClass={cs.CV},
      url={https://arxiv.org/abs/2001.08499}, 
}

@InProceedings{SSM,
    author    = {Zubic, Nikola and Gehrig, Mathias and Scaramuzza, Davide},
    title     = {State Space Models for Event Cameras},
    booktitle = {Proceedings of the IEEE/CVF Conference on Computer Vision and Pattern Recognition (CVPR)},
    month     = {June},
    year      = {2024},
    pages     = {5819-5828}
}

@InProceedings{Owl-vit,
author="Minderer, Matthias
and Gritsenko, Alexey
and Stone, Austin
and Neumann, Maxim
and Weissenborn, Dirk
and Dosovitskiy, Alexey
and Mahendran, Aravindh
and Arnab, Anurag
and Dehghani, Mostafa
and Shen, Zhuoran
and Wang, Xiao
and Zhai, Xiaohua
and Kipf, Thomas
and Houlsby, Neil",
editor="Avidan, Shai
and Brostow, Gabriel
and Ciss{\'e}, Moustapha
and Farinella, Giovanni Maria
and Hassner, Tal",
title="Simple Open-Vocabulary Object Detection",
booktitle="Computer Vision -- ECCV 2022",
year="2022",
publisher="Springer Nature Switzerland",
address="Cham",
pages="728--755",
isbn="978-3-031-20080-9"
}

@InProceedings{ground_dino,
author="Liu, Shilong
and Zeng, Zhaoyang
and Ren, Tianhe
and Li, Feng
and Zhang, Hao
and Yang, Jie
and Jiang, Qing
and Li, Chunyuan
and Yang, Jianwei
and Su, Hang
and Zhu, Jun
and Zhang, Lei",
editor="Leonardis, Ale{\v{s}}
and Ricci, Elisa
and Roth, Stefan
and Russakovsky, Olga
and Sattler, Torsten
and Varol, G{\"u}l",
title="Grounding DINO: Marrying DINO with Grounded Pre-training for Open-Set Object Detection",
booktitle="Computer Vision -- ECCV 2024",
year="2025",
publisher="Springer Nature Switzerland",
address="Cham",
pages="38--55",
isbn="978-3-031-72970-6"
}

@ARTICLE{DEOE,
  author={Zhang, Haitian and Xu, Chang and Wang, Xinya and Liu, Bingde and Hua, Guang and Yu, Lei and Yang, Wen},
  journal={IEEE Transactions on Pattern Analysis and Machine Intelligence}, 
  title={Detecting Every Object From Events}, 
  year={2025},
  volume={47},
  number={8},
  pages={7171-7178},
  doi={10.1109/TPAMI.2025.3565102}}

@InProceedings{Cmax1,
author = {Gallego, Guillermo and Rebecq, Henri and Scaramuzza, Davide},
title = {A Unifying Contrast Maximization Framework for Event Cameras, With Applications to Motion, Depth, and Optical Flow Estimation},
booktitle = {Proceedings of the IEEE Conference on Computer Vision and Pattern Recognition (CVPR)},
month = {June},
year = {2018}
}

@InProceedings{Cmax2,
author = {Stoffregen, Timo and Kleeman, Lindsay},
title = {Event Cameras, Contrast Maximization and Reward Functions: An Analysis},
booktitle = {Proceedings of the IEEE/CVF Conference on Computer Vision and Pattern Recognition (CVPR)},
month = {June},
year = {2019}
}

@ARTICLE{graph-cuts,
  author={Zhou, Yi and Gallego, Guillermo and Lu, Xiuyuan and Liu, Siqi and Shen, Shaojie},
  journal={IEEE Transactions on Neural Networks and Learning Systems}, 
  title={Event-Based Motion Segmentation With Spatio-Temporal Graph Cuts}, 
  year={2023},
  volume={34},
  number={8},
  pages={4868-4880},
  doi={10.1109/TNNLS.2021.3124580}}

@InProceedings{stoffregen,
author = {Stoffregen, Timo and Gallego, Guillermo and Drummond, Tom and Kleeman, Lindsay and Scaramuzza, Davide},
title = {Event-Based Motion Segmentation by Motion Compensation},
booktitle = {Proceedings of the IEEE/CVF International Conference on Computer Vision (ICCV)},
month = {October},
year = {2019}
}

@INPROCEEDINGS{cascaded,
  author={Lu, Xiuyuan and Zhou, Yi and Shen, Shaojie},
  booktitle={2021 IEEE/RSJ International Conference on Intelligent Robots and Systems (IROS)}, 
  title={Event-based Motion Segmentation by Cascaded Two-Level Multi-Model Fitting}, 
  year={2021},
  volume={},
  number={},
  pages={4445-4452},
  doi={10.1109/IROS51168.2021.9636307}}

@inproceedings{shiba,
  title={Simultaneous motion and noise estimation with event cameras},
  author={Shiba, Shintaro and Aoki, Yoshimitsu and Gallego, Guillermo},
  booktitle={Proceedings of the IEEE/CVF International Conference on Computer Vision},
  pages={6959--6969},
  year={2025}
}

@inproceedings{STCF,
  title={Design of a spatiotemporal correlation filter for event-based sensors},
  author={Liu, Hongjie and Brandli, Christian and Li, Chenghan and Liu, Shih-Chii and Delbruck, Tobi},
  booktitle={2015 IEEE International Symposium on Circuits and Systems (ISCAS)},
  pages={722--725},
  year={2015},
  organization={IEEE}
}

@ARTICLE{knoise,
  author={Khodamoradi, Alireza and Kastner, Ryan},
  journal={IEEE Transactions on Emerging Topics in Computing}, 
  title={$O(N)$O(N)-Space Spatiotemporal Filter for Reducing Noise in Neuromorphic Vision Sensors}, 
  year={2021},
  volume={9},
  number={1},
  pages={15-23},
  doi={10.1109/TETC.2017.2788865}}

@InProceedings{EDnCNN,
author = {Baldwin, R. Wes and Almatrafi, Mohammed and Asari, Vijayan and Hirakawa, Keigo},
title = {Event Probability Mask (EPM) and Event Denoising Convolutional Neural Network (EDnCNN) for Neuromorphic Cameras},
booktitle = {Proceedings of the IEEE/CVF Conference on Computer Vision and Pattern Recognition (CVPR)},
month = {June},
year = {2020}
}

@inproceedings{eventzoom,
  title={EventZoom: Learning to denoise and super resolve neuromorphic events},
  author={Duan, Peiqi and Wang, Zihao W and Zhou, Xinyu and Ma, Yi and Shi, Boxin},
  booktitle={Proceedings of the IEEE/CVF conference on computer vision and pattern recognition},
  pages={12824--12833},
  year={2021}
}

@article{emlb,
  title={E-MLB: Multilevel benchmark for event-based camera denoising},
  author={Ding, Saizhe and Chen, Jinze and Wang, Yang and Kang, Yu and Song, Weiguo and Cheng, Jie and Cao, Yang},
  journal={IEEE Transactions on Multimedia},
  volume={26},
  pages={65--76},
  year={2023},
  publisher={IEEE}
}

@inproceedings{DBSCAN1,
  author       = {Martin Ester and
                  Hans{-}Peter Kriegel and
                  J{\"{o}}rg Sander and
                  Xiaowei Xu},
  editor       = {Evangelos Simoudis and
                  Jiawei Han and
                  Usama M. Fayyad},
  title        = {A Density-Based Algorithm for Discovering Clusters in Large Spatial
                  Databases with Noise},
  booktitle    = {Proceedings of the Second International Conference on Knowledge Discovery
                  and Data Mining (KDD-96), Portland, Oregon, {USA}},
  pages        = {226--231},
  publisher    = {{AAAI} Press},
  year         = {1996},
  url          = {http://www.aaai.org/Library/KDD/1996/kdd96-037.php},
  bibsource    = {dblp computer science bibliography, https://dblp.org}
}

@article{DBSCAN2,
author = {Davide Falanga  and Kevin Kleber  and Davide Scaramuzza },
title = {Dynamic obstacle avoidance for quadrotors with event cameras},
journal = {Science Robotics},
volume = {5},
number = {40},
pages = {eaaz9712},
year = {2020},
doi = {10.1126/scirobotics.aaz9712},
URL = {https://www.science.org/doi/abs/10.1126/scirobotics.aaz9712},
eprint = {https://www.science.org/doi/pdf/10.1126/scirobotics.aaz9712}}

@ARTICLE{DBSCAN3,
  author={Iaboni, Craig and Patel, Himanshu and Lobo, Deepan and Choi, Ji-Won and Abichandani, Pramod},
  journal={IEEE Access}, 
  title={Event Camera Based Real-Time Detection and Tracking of Indoor Ground Robots}, 
  year={2021},
  volume={9},
  number={},
  pages={166588-166602},
  doi={10.1109/ACCESS.2021.3133533}}

@misc{DBSCAN4,
      title={Speed-based Filtration and DBSCAN of Event-based Camera Data with Neuromorphic Computing}, 
      author={Charles P. Rizzo and Catherine D. Schuman and James S. Plank},
      year={2024},
      eprint={2401.15212},
      archivePrefix={arXiv},
      primaryClass={cs.NE},
      url={https://arxiv.org/abs/2401.15212}, 
}

@misc{DBSCAN5,
      title={A Neuromorphic Implementation of the DBSCAN Algorithm}, 
      author={Charles P. Rizzo and James S. Plank},
      year={2024},
      eprint={2409.14298},
      archivePrefix={arXiv},
      primaryClass={cs.NE},
      url={https://arxiv.org/abs/2409.14298}, 
}

@InProceedings{HDBSCAN,
author="Campello, Ricardo J. G. B.
and Moulavi, Davoud
and Sander, Joerg",
editor="Pei, Jian
and Tseng, Vincent S.
and Cao, Longbing
and Motoda, Hiroshi
and Xu, Guandong",
title="Density-Based Clustering Based on Hierarchical Density Estimates",
booktitle="Advances in Knowledge Discovery and Data Mining",
year="2013",
publisher="Springer Berlin Heidelberg",
address="Berlin, Heidelberg",
pages="160--172",
isbn="978-3-642-37456-2"
}

@book{morton1966computer,
  title={A computer oriented geodetic data base and a new technique in file sequencing},
  author={Morton, Guy M},
  year={1966},
  publisher={International Business Machines Company}
}

@ARTICLE{morton2,
  author={Connor, Michael and Kumar, Piyush},
  journal={IEEE Transactions on Visualization and Computer Graphics}, 
  title={Fast construction of k-nearest neighbor graphs for point clouds}, 
  year={2010},
  volume={16},
  number={4},
  pages={599-608},
  doi={10.1109/TVCG.2010.9}}

@inproceedings{orenstein,
author = {Orenstein, J. A. and Merrett, T. H.},
title = {A class of data structures for associative searching},
year = {1984},
isbn = {0897911288},
publisher = {Association for Computing Machinery},
address = {New York, NY, USA},
url = {https://doi.org/10.1145/588011.588037},
doi = {10.1145/588011.588037},
booktitle = {Proceedings of the 3rd ACM SIGACT-SIGMOD Symposium on Principles of Database Systems},
pages = {181–190},
numpages = {10},
location = {Waterloo, Ontario, Canada},
series = {PODS '84}
}

@inproceedings{pointcloud1,
author = {Liu, Xueyuan and Song, Zhuoran and Chen, Hao and Li, Xing and Liang, Xiaoyao},
title = {MoC: A Morton-Code-Based Fine-Grained Quantization for Accelerating Point Cloud Neural Networks},
year = {2024},
isbn = {9798400706011},
publisher = {Association for Computing Machinery},
address = {New York, NY, USA},
url = {https://doi.org/10.1145/3649329.3655905},
doi = {10.1145/3649329.3655905},
booktitle = {Proceedings of the 61st ACM/IEEE Design Automation Conference},
articleno = {42},
numpages = {6},
location = {San Francisco, CA, USA},
series = {DAC '24}
}

@INPROCEEDINGS{pointcloud2,
  author={Kim, Taehoon and Kim, Kyoung-Sook and Lee, Jun and Matono, Akiyoshi and Li, Ki-Joune},
  booktitle={2019 IEEE International Conference on Big Data and Smart Computing (BigComp)}, 
  title={Efficient Encoding and Decoding Extended Geocodes for Massive Point Cloud Data}, 
  year={2019},
  volume={},
  number={},
  pages={1-8},
  doi={10.1109/BIGCOMP.2019.8679177}}

@inproceedings{pointcloud3,
  title={Parallel processing of big point clouds using Z-Order-based partitioning},
  author={Alis, C and Boehm, J and Liu, K},
  booktitle={International Archives of the Photogrammetry, Remote Sensing and Spatial Information Sciences-ISPRS Archives},
  volume={41},
  pages={71--77},
  year={2016},
  organization={International Society of Photogrammetry and Remote Sensing (ISPRS)}
}

@INPROCEEDINGS{fixed_thresh1,
  author={Xiao, Kanglin and Cui, Xiaoxin and Liu, Kefei and Cui, Xiaole and Wang, Xin'an},
  booktitle={2021 International Joint Conference on Neural Networks (IJCNN)}, 
  title={An SNN-Based and Neuromorphic-Hardware-Implementable Noise Filter with Self-adaptive Time Window for Event-Based Vision Sensor}, 
  year={2021},
  volume={},
  number={},
  pages={1-8},
  doi={10.1109/IJCNN52387.2021.9534073}}

@ARTICLE{fixed_thresh2,
  author={Fang, Huachen and Wu, Jinjian and Hou, Qibin and Dong, Weisheng and Shi, Guangming},
  journal={IEEE Transactions on Pattern Analysis and Machine Intelligence}, 
  title={Fast Window-Based Event Denoising With Spatiotemporal Correlation Enhancement}, 
  year={2025},
  volume={47},
  number={3},
  pages={1381-1394},
  doi={10.1109/TPAMI.2024.3467709}}

@article{rejectionsampling,
  title={13. various techniques used in connection with random digits},
  author={Von Neumann, John},
  journal={Appl. Math Ser},
  volume={12},
  number={36-38},
  pages={3},
  year={1951}
}

@INPROCEEDINGS{barranco,
  author={Barranco, Francisco and Fermuller, Cornelia and Ros, Eduardo},
  booktitle={2018 IEEE/RSJ International Conference on Intelligent Robots and Systems (IROS)}, 
  title={Real-Time Clustering and Multi-Target Tracking Using Event-Based Sensors}, 
  year={2018},
  volume={},
  number={},
  pages={5764-5769},
  doi={10.1109/IROS.2018.8593380}}

@InProceedings{GSCEventMOD,
    author    = {Mondal, Anindya and R, Shashant and Giraldo, Jhony H. and Bouwmans, Thierry and Chowdhury, Ananda S.},
    title     = {Moving Object Detection for Event-Based Vision Using Graph Spectral Clustering},
    booktitle = {Proceedings of the IEEE/CVF International Conference on Computer Vision (ICCV) Workshops},
    month     = {October},
    year      = {2021},
    pages     = {876-884}
}

@INPROCEEDINGS{MondalKmeans,
  author={Mondal, Anindya and Das, Mayukhmali},
  booktitle={2021 IEEE 8th Uttar Pradesh Section International Conference on Electrical, Electronics and Computer Engineering (UPCON)}, 
  title={Moving Object Detection for Event-based Vision using k-means Clustering}, 
  year={2021},
  volume={},
  number={},
  pages={1-6},
  doi={10.1109/UPCON52273.2021.9667636}}

@Article{z-order1,
author={Yang, HuiJun
and Chang, Jian
and Geng, Nan
and Notman, Gabriel
and Li, Shuqin
and Jiang, Min
and Wang, MeiLi
and Zhang, JianJun},
title={Texture organisation and mapping on Citrus sinensis point cloud},
journal={Multimedia Tools and Applications},
year={2017},
month={Jul},
day={01},
volume={76}
}

@Article{z-order2,
title = {Binarized-octree generation for Cartesian adaptive mesh refinement around immersed geometries},
journal = {Journal of Computational Physics},
volume = {368},
pages = {179-195},
year = {2018},
issn = {0021-9991},
doi = {https://doi.org/10.1016/j.jcp.2018.04.039},
url = {https://www.sciencedirect.com/science/article/pii/S002199911830264X},
author = {Jaber J. Hasbestan and Inanc Senocak}
}

@Inbook{z-order3,
author="Haverkort, Herman
and Toma, Laura",
title="Quadtrees and Morton Indexing",
bookTitle="Encyclopedia of Algorithms",
year="2016",
publisher="Springer New York",
address="New York, NY",
pages="1637--1642",
isbn="978-1-4939-2864-4",
doi="10.1007/978-1-4939-2864-4_585",
url="https://doi.org/10.1007/978-1-4939-2864-4_585"
}

@inproceedings{karras2012maximizing,
  title={Maximizing parallelism in the construction of BVHs, octrees, and k-d trees},
  author={Karras, Tero},
  booktitle={Proceedings of the Fourth ACM SIGGRAPH/Eurographics Conference on High-Performance Graphics},
  pages={33--37},
  year={2012}
}

@article{magrini2025fred,
  title={FRED: The Florence RGB-Event Drone Dataset},
  author={Magrini, Gabriele and Marini, Niccol{\`o} and Becattini, Federico and Berlincioni, Lorenzo and Biondi, Niccol{\`o} and Pala, Pietro and Del Bimbo, Alberto},
  journal={arXiv preprint arXiv:2506.05163},
  year={2025}
}

@InProceedings{etram,
    author    = {Verma, Aayush Atul and Chakravarthi, Bharatesh and Vaghela, Arpitsinh and Wei, Hua and Yang, Yezhou},
    title     = {eTraM: Event-based Traffic Monitoring Dataset},
    booktitle = {Proceedings of the IEEE/CVF Conference on Computer Vision and Pattern Recognition (CVPR)},
    month     = {June},
    year      = {2024},
    pages     = {22637-22646}
}

@article{ST-DBSCAN,
title = {ST-DBSCAN: An algorithm for clustering spatial–temporal data},
journal = {Data \& Knowledge Engineering},
volume = {60},
number = {1},
pages = {208-221},
year = {2007},
note = {Intelligent Data Mining},
issn = {0169-023X},
doi = {https://doi.org/10.1016/j.datak.2006.01.013},
url = {https://www.sciencedirect.com/science/article/pii/S0169023X06000218},
author = {Derya Birant and Alp Kut}
}

@article{meanshift,
  title={Mean shift: A robust approach toward feature space analysis},
  author={Comaniciu, Dorin and Meer, Peter},
  journal={IEEE Transactions on pattern analysis and machine intelligence},
  volume={24},
  number={5},
  pages={603--619},
  year={2002},
  publisher={IEEE}
}

@article{mcinnes2017hdbscan,
  title={hdbscan: Hierarchical density based clustering.},
  author={McInnes, Leland and Healy, John and Astels, Steve and others},
  journal={J. Open Source Softw.},
  volume={2},
  number={11},
  pages={205},
  year={2017}
}

@InProceedings{evgait,
author = {Wang, Yanxiang and Du, Bowen and Shen, Yiran and Wu, Kai and Zhao, Guangrong and Sun, Jianguo and Wen, Hongkai},
title = {EV-Gait: Event-Based Robust Gait Recognition Using Dynamic Vision Sensors},
booktitle = {Proceedings of the IEEE/CVF Conference on Computer Vision and Pattern Recognition (CVPR)},
month = {June},
year = {2019}
}

@inproceedings{BAF,
  title={Frame-free dynamic digital vision},
  author={Delbruck, Tobi and others},
  booktitle={Proceedings of Intl. Symp. on Secure-Life Electronics, Advanced Electronics for Quality Life and Society},
  volume={1},
  pages={21--26},
  year={2008},
  organization={Tokyo}
}

@ARTICLE{Lagorce,
  author={Lagorce, Xavier and Orchard, Garrick and Galluppi, Francesco and Shi, Bertram E. and Benosman, Ryad B.},
  journal={IEEE Transactions on Pattern Analysis and Machine Intelligence}, 
  title={HOTS: A Hierarchy of Event-Based Time-Surfaces for Pattern Recognition}, 
  year={2017},
  volume={39},
  number={7},
  pages={1346-1359},
  doi={10.1109/TPAMI.2016.2574707}}

@Article{ynoise,
AUTHOR = {Feng, Yang and Lv, Hengyi and Liu, Hailong and Zhang, Yisa and Xiao, Yuyao and Han, Chengshan},
TITLE = {Event Density Based Denoising Method for Dynamic Vision Sensor},
JOURNAL = {Applied Sciences},
VOLUME = {10},
YEAR = {2020},
NUMBER = {6},
ARTICLE-NUMBER = {2024},
URL = {https://www.mdpi.com/2076-3417/10/6/2024},
ISSN = {2076-3417},
DOI = {10.3390/app10062024}
}

@InProceedings{guo,
    author    = {Rios-Navarro, Antonio and Guo, Shasha and Gnaneswaran, Abarajithan and Vijayakumar, Keerthivasan and Linares-Barranco, Alejandro and Aarrestad, Thea and Kastner, Ryan and Delbruck, Tobi},
    title     = {Within-Camera Multilayer Perceptron DVS Denoising},
    booktitle = {Proceedings of the IEEE/CVF Conference on Computer Vision and Pattern Recognition (CVPR) Workshops},
    month     = {June},
    year      = {2023},
    pages     = {3933-3942}
}

@ARTICLE{dwf,
  author={Guo, Shasha and Delbruck, Tobi},
  journal={IEEE Transactions on Pattern Analysis and Machine Intelligence}, 
  title={Low Cost and Latency Event Camera Background Activity Denoising}, 
  year={2023},
  volume={45},
  number={1},
  pages={785-795},
  doi={10.1109/TPAMI.2022.3152999}}

@InProceedings{simultaneous,
    author    = {Shiba, Shintaro and Aoki, Yoshimitsu and Gallego, Guillermo},
    title     = {Simultaneous Motion And Noise Estimation with Event Cameras},
    booktitle = {Proceedings of the IEEE/CVF International Conference on Computer Vision (ICCV)},
    month     = {October},
    year      = {2025},
    pages     = {6959-6969}
}

@ARTICLE{GEF,
  author={Duan, Peiqi and Wang, Zihao W. and Shi, Boxin and Cossairt, Oliver and Huang, Tiejun and Katsaggelos, Aggelos K.},
  journal={IEEE Transactions on Pattern Analysis and Machine Intelligence}, 
  title={Guided Event Filtering: Synergy Between Intensity Images and Neuromorphic Events for High Performance Imaging}, 
  year={2022},
  volume={44},
  number={11},
  pages={8261-8275},
  doi={10.1109/TPAMI.2021.3113344}}

@InProceedings{Baldwin,
author="Baldwin, R. Wes
and Almatrafi, Mohammed
and Kaufman, Jason R.
and Asari, Vijayan
and Hirakawa, Keigo",
editor="Karray, Fakhri
and Campilho, Aur{\'e}lio
and Yu, Alfred",
title="Inceptive Event Time-Surfaces for Object Classification Using Neuromorphic Cameras",
booktitle="Image Analysis and Recognition",
year="2019",
publisher="Springer International Publishing",
address="Cham",
pages="395--403",
isbn="978-3-030-27272-2"
}

@InProceedings{edformer,
author="Jiang, Bin
and Xiong, Bo
and Qu, Bohan
and Salman Asif, M.
and Zhou, You
and Ma, Zhan",
editor="Leonardis, Ale{\v{s}}
and Ricci, Elisa
and Roth, Stefan
and Russakovsky, Olga
and Sattler, Torsten
and Varol, G{\"u}l",
title="EDformer: Transformer-Based Event Denoising Across Varied Noise Levels",
booktitle="Computer Vision -- ECCV 2024",
year="2025",
publisher="Springer Nature Switzerland",
address="Cham",
pages="200--216",
isbn="978-3-031-73383-3"
}

@article{Ding,
author = {Ding, Saizhe and Zhang, Haorui and Zhang, Yuxin and Huang, Xinyan and Song, Weiguo},
title = {Hyper real-time flame detection: Dynamic insights from event cameras and FlaDE dataset},
year = {2025},
issue_date = {Mar 2025},
publisher = {Pergamon Press, Inc.},
address = {USA},
volume = {263},
number = {C},
issn = {0957-4174},
url = {https://doi.org/10.1016/j.eswa.2024.125746},
doi = {10.1016/j.eswa.2024.125746},
journal = {Expert Syst. Appl.},
month = mar,
numpages = {14}
}

@INPROCEEDINGS{zhao,
  author={Zhao, Chunhui and Li, Yakun and Lyu, Yang},
  booktitle={2023 IEEE International Conference on Robotics and Automation (ICRA)}, 
  title={Event-based Real-time Moving Object Detection Based On IMU Ego-motion Compensation}, 
  year={2023},
  volume={},
  number={},
  pages={690-696},
  doi={10.1109/ICRA48891.2023.10160472}}

@INPROCEEDINGS{zhu,
  author={Zhu, Alex Zihao and Yuan, Liangzhe and Chaney, Kenneth and Daniilidis, Kostas},
  booktitle={2019 IEEE/CVF Conference on Computer Vision and Pattern Recognition Workshops (CVPRW)}, 
  title={Live Demonstration: Unsupervised Event-Based Learning of Optical Flow, Depth and Egomotion}, 
  year={2019},
  volume={},
  number={},
  pages={1694-1694},
  doi={10.1109/CVPRW.2019.00216}}

@inproceedings{Optuna,
  title={Optuna: A next-generation hyperparameter optimization framework},
  author={Akiba, Takuya and Sano, Shotaro and Yanase, Toshihiko and Ohta, Takeru and Koyama, Masanori},
  booktitle={Proceedings of the 25th ACM SIGKDD international conference on knowledge discovery \& data mining},
  pages={2623--2631},
  year={2019}
}

@article{TPE,
  title={Tree-structured parzen estimator: Understanding its algorithm components and their roles for better empirical performance},
  author={Watanabe, Shuhei},
  journal={arXiv preprint arXiv:2304.11127},
  year={2023}
}

\end{document}